\documentclass[10pt,twocolumn,letterpaper]{article}

\usepackage[pagenumbers]{cvpr} 

\definecolor{cvprblue}{rgb}{0.21,0.49,0.74}
\usepackage[pagebackref,breaklinks,colorlinks,citecolor=cvprblue]{hyperref}
\usepackage{multirow}
\usepackage[ruled,lined,noend]{algorithm2e}
\usepackage{colortbl}

\def\paperID{3436} 
\def\confName{CVPR}
\def\confYear{2025}

\title{Modality Alignment Meets Federated Broadcasting}

\author{Yuting Ma\\
University of Science and Technology of China\\
{\tt\small ytma@ustc.edu.cn}
\and
Shengeng Tang\\
Hefei University of Technology\\
{\tt\small tangsg@hfut.edu.cn}
\and
Xiaohua Xu\\
University of Science and Technology of China\\
{\tt\small xiaohuaxu@ustc.edu.cn}
\and
Lechao Cheng$^*$\\
Hefei University of Technology\\
{\tt\small chenglc@hfut.edu.cn}
}

\begin{document}
\maketitle
\begin{abstract}
Federated learning (FL) has emerged as a powerful approach to safeguard data privacy by training models across distributed edge devices without centralizing local data. Despite advancements in homogeneous data scenarios, maintaining performance between the global and local clients in FL over heterogeneous data remains challenging due to data distribution variations that degrade model convergence and increase computational costs. This paper introduces a novel FL framework leveraging modality alignment, where a text encoder resides on the server, and image encoders operate on local devices. Inspired by multi-modal learning paradigms like CLIP, this design aligns cross-client learning by treating server-client communications akin to multi-modal broadcasting. We initialize with a pre-trained model to mitigate overfitting, updating select parameters through low-rank adaptation (LoRA) to meet computational demand and performance efficiency. Local models train independently and communicate updates to the server, which aggregates parameters via a query-based method, facilitating cross-client knowledge sharing and performance improvement under extreme heterogeneity. Extensive experiments on benchmark datasets demonstrate the framework’s efficacy in maintaining generalization and robustness, even in highly heterogeneous settings.

\end{abstract}  
\section{Introduction}

Federated learning \cite{FedAVG} is an advanced distributed machine learning paradigm that aims to protect data privacy by collaboratively training a global model across distributed clients without the need to centralize local data. In this framework, each edge device independently performs local training using its private data and transmits only model updates, such as weight parameters or gradients, to a central server for aggregation. Subsequently, the server derives global updates and then broadcasts them back to clients, facilitating iterative refinement while maintaining data locality. 
Consequently, federated learning has shown exceptional promise in privacy-sensitive applications, particularly in sectors like finance \cite{FL1, FL2} and healthcare \cite{FL3, FL4}, where stringent data confidentiality regulations and security requirements are paramount.
\begin{figure}
    \centering
    \includegraphics[width=\linewidth]{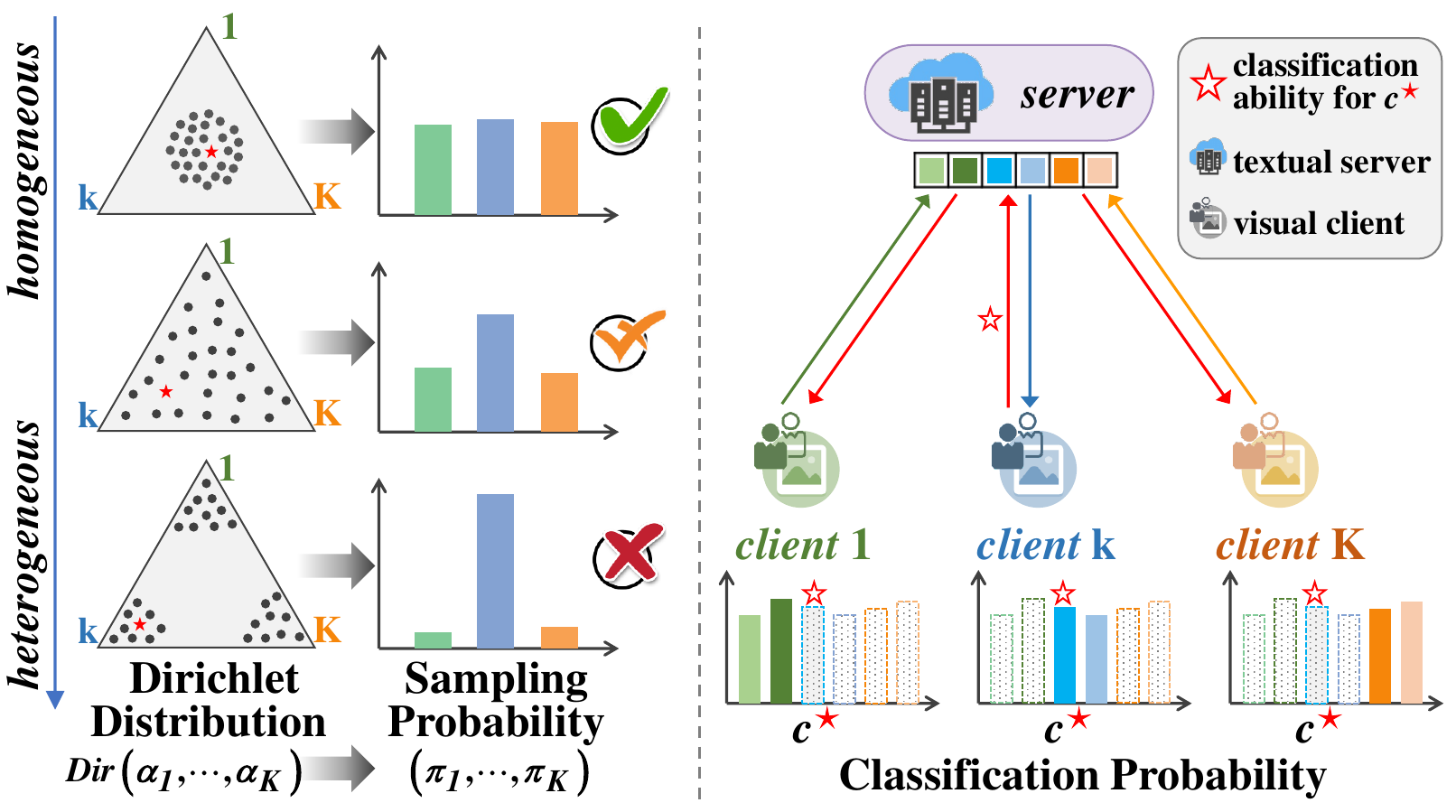}
    \caption{Left: Data distribution approximation based on Dirichlet distribution. Right: By leveraging a global text module on the server as a “constraint”, we can anchor image category learning across distributed nodes to achieve knowledge sharing.}
    \vspace{-5mm}
    \label{fig:teaser}
\end{figure}

Advancements \cite{FedProx, SCAFFOLD, FedDyn, Moon, FedBABU, FedRep, pFL, FedPAC, FedHPL} in FL have made considerable efforts in the heterogeneous data setting to adapt to real-world scenes. For example, FedProx \cite{FedProx} applies a local regularization term to adapt the model drift, contributing to model stability and training convergence. pFL \cite{pFL} has achieved notable success in heterogeneous data environments by allowing each client to train a personalized model that adapts to their own data distribution. While these traditional approaches attempt to meet the heterogeneous data challenge by optimizing local optimization objectives and refining aggregation processes, they often suffer from slow convergence and increased computational costs. Furthermore, these methods struggle to maintain the efficiency of both global and local performance in extreme heterogeneity scenarios. These challenges highlight the difficulty of achieving consistent and reliable performance across diverse federated learning environments, which we address in greater detail in the following sections.


Let us delve into the inherent complexity of data heterogeneity. In real-world applications, data heterogeneity is often an opaque black-box, making it difficult to define or predict precisely. To model this heterogeneity, researchers commonly use imbalanced sampling, \eg Dirichlet strategy \cite{Dirichlet} with a concentration parameter $\alpha$. As $\alpha$ becomes smaller, the distribution becomes more imbalanced, resulting in a pronounced class separation, as illustrated in Figure~\ref{fig:teaser}. Existing methods have made progress in improving local model performance for heterogeneous data scenarios. However, extreme heterogeneity (\eg severe class separation) leads to severe overfitting, resulting in poor global model performance and generalization. It remains a critical challenge in FL and is increasingly relevant in real-world scenarios, where data isolation plays a key role in ensuring privacy protections.


Achieving effective local updating and maintaining global generalization in such heterogeneous scenarios seems highly complex. However, drawing inspiration from the decoupled modality CLIP \cite{CLIP}, we observe the key similarity in multimodal learning with FL: 1) the optimization of the image branch corresponds to the local updating of edge devices; 2) the process of aligning one modality (\eg, text) with another (\eg, image) bears striking parallels to server-to-client broadcasting in FL. This insight motivates us to explore whether a server-based text module can be deployed to align with client-based image modules, facilitating global generalization and local optimization. The advantages of this approach are evident: by leveraging a global text module on the server as a “constraint”, we can anchor image category learning across distributed nodes. This mechanism enables effective guidance of federated aggregation, even in the presence of extreme data heterogeneity and class separation. Although some recent studies \cite{FedOTP, FedPGP} have explored multimodal strategies, their focus has largely been on harnessing the strong representations of pre-trained multimodal models at the edge, rather than aligning multimodal information for FL.

In this paper, we introduce a novel framework called FedAlign in heterogeneous federated learning that establishes a bridge between the central server and local clients. We make an early attempt to implement modality alignment by decomposing a pre-trained CLIP model into two distinct components: a text encoder issued to the server and an image encoder deployed on clients. The text encoder is globally trained by using knowledge uploaded from clients, while the image encoder undergoes local training with private data. In particular, training models from scratch on limited data often leads to slow convergence, massive computations, and potential overfitting. To mitigate these issues, we initialize the model with a pre-trained model (PM) that possesses strong representations and then update only a small set of additional model parameters. Specifically, we fine-tune the frozen PM by using the LoRA strategy with its additive property instead of prompt tuning to better adapt both global and local performance without inference latency. In this framework, clients perform local training and upload image feature embeddings with updated local parameters to the server. This facilitates the alignment of global and local training processes and ensures consistency in visual-language modality matching. Following federated learning protocols, the server aggregates and broadcasts updates from clients, thereby enabling cross-client knowledge interaction. To further improve adaptability in various data scenarios, we propose a query-based parameter aggregation method that effectively balances class-specific representations with the number of available samples, ensuring robust performance even under extreme heterogeneous conditions.
In summary, the contributions are as follows.

\begin{itemize}
    \item We apply modality alignment in FL to bridge server-client broadcast. By deploying the text encoder on the server and image encoders on clients, we enable cross-client knowledge sharing and effective local optimization.
    \item We propose a parameter aggregation method tailored to federated tasks, incorporating efficient parameter tuning strategies to balance model generalization and personalization objectives effectively.
    \item We perform extensive validation on commonly public datasets, with experimental results that demonstrate the superiority of our method and its reliability in handling extreme data heterogeneity.
\end{itemize}

\section{Related Works}
\subsection{Pre-trained Model in Federated Learning}
With the development of PMs \cite{ViT, CLIP}, incorporating robust representations and generalization capabilities into parameter fine-tuning for model performance improvement has gained increasing attention.
Recently, some FL efforts have started to investigate the above effectiveness mechanism and get remarkable improvement.
For instance, \cite{ViT_FL} employs the pre-trained Vision Transformer (ViT) as an initialized global model and updates the entire model locally.
However, the full fine-tuning algorithm increases the computation and communication costs, causing limited performance improvement.
To tackle this challenge, some studies \cite{pFedPT, pFedPG, SGPT, DAM_VP} utilize prompt learning \cite{VPT} in downstream tasks, which considers the PM as the frozen backbone and only optimizes a few learnable task-specific vectors.
In contrast to vision-based model architectures, \cite{PromptFL, FedOTP, FedPGP} apply a pre-trained vision-language model (\ie CLIP) as the frozen backbone and update the textual prompts to adapt data heterogeneity.
Compared to these methods, our framework maintains both global and personalized performances by decoupling training and parameter aggregation.

\subsection{Parameter Efficient Fine-tuning in Vision-Language Model}
Vision-language models (VLMs), like CLIP \cite{CLIP}, leverage a self-supervised paradigm with massive web-scale data for training. Thus, well-trained VLMs contain rich multimodal representations. 
However, the well-trained VLM still needs to employ fine-tuning to effectively adapt to specific downstream tasks.
Various approaches, such as linear probing \cite{LP1}, adapter-based method \cite{AD1, AD2}, and prompt-based algorithm \cite{PT1, PT2, PT3} have been proposed to achieve fine-tuning.
Among them, prompt learning is a particularly popular research direction.
Initially, the handcrafted instructions (\eg `a photo of a [classname]') are considered as text prompts into the language branch of VLM, allowing it to understand the specific task.
Subsequently, CoOp \cite{CoOp} introduces continuous prompt vectors at its language branch for a downstream task and automatically learns prompts during the fine-tuning stage. Co-CoOp \cite{CoCoOp} further improves the generalization of CoOp by explicitly conditioning prompts on images.
MaPLe \cite{MaPLe} extends the single-modal prompting to the multimodal design by improving strong prompt coupling and representation alignment between vision and language branches.
Recently, a technique called LoRA \cite{LoRA, CLIP_LoRA, FLoRA} has been utilized to eliminate the extra computational requirement of the inference stage. It adds low-rank matrices to explicitly represent model weight changes while keeping the original model parameters frozen.
In FedAlign, we employ LoRA to update encoder parameters of vision and language branches in the FL community.

\section{Preliminary}\label{sec:pre}
We first give a brief introduction to the FL quagmire in severe data heterogeneity and then revisit the modality alignment process which comprises text and image encoding. 

\subsection{Quagmire in Data Heterogeneity}
It is noteworthy that recent studies \cite{FedPL, FedFed, FuseFL, FedPCL, FPL} often model data heterogeneity in federated learning by parameterizing and constructing a Dirichlet distribution. Therefore, we first review the related definition in a federated system with $K$ clients and $C$ classes.
For each class $c$, we define the sampling probability on edge devices as $Dir(\alpha_1$, ..., $\alpha_K)$\footnote{Here we omit class superscript $c$ for simplicity.}. Consequently, the Dirichlet probability density function is defined as follows:
\begin{equation}
\begin{split}
     Dir(\alpha_1, ..., \alpha_K) & \triangleq \frac{\Gamma\left(\sum_{k=1}^K \alpha_k\right)}{\prod_{k=1}^K \Gamma(\alpha_k)} \prod_{k=1}^K \pi_k^{\alpha_k - 1},
\end{split}
\end{equation}
where $\Gamma(.)$ denotes the Gamma function and each $\alpha_k > 0$. In addition, $\pi_k$ represents the sampling probability of the class $c$ assigned to the client $k$. Therefore, each $\pi_k \geq 0$ and $\sum_{k=1}^{K}\pi_k = 1$. Technically, $(\pi_1, ..., \pi_K)$ indicates a $K-1$ dimensional probability simplex in a $K$ dimensional space.

\begin{figure}
    \centering
    \includegraphics[width=0.9\linewidth, trim=4cm 1cm 4cm 2cm, clip]{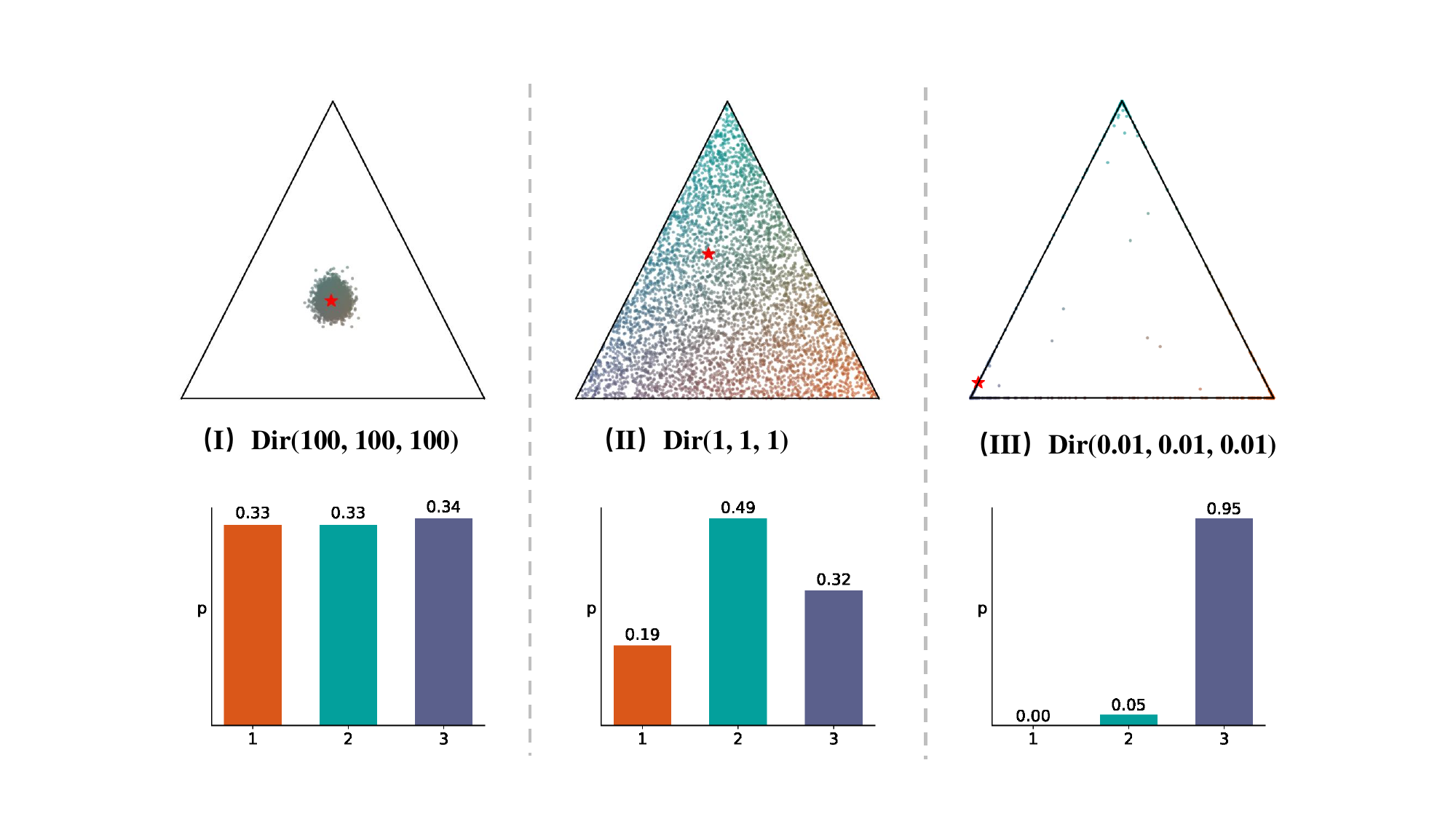}
    \vspace{-4mm}
    \caption{(I) When $\alpha$ is set to a large value (\eg, $\alpha=100$), the probabilities focus mainly on near-uniform distributions. (II) When $\alpha = 1$, the imbalanced distribution uniformly exists in $K$ dimensional space. (III) When $\alpha$ approaches $0$, the probability over clients degenerates into a near one-hot partition.}
    \vspace{-5mm}
    \label{fig:dir}
\end{figure}

\noindent\textbf{Remark.} For simplicity, researchers often adopt a symmetric Dirichlet distribution in FL, that is, each parameter $\alpha_k$ has the same value. In this case, the density with $K$ clients shares a symmetric distribution ($Dir(\alpha)$, $\alpha_1 = \alpha_2 = ... = \alpha_K = \alpha$). We will take this setting to facilitate further explorations, as shown in Figure~\ref{fig:dir}. As the parameter $\alpha$ evolves from large to small, the simulated data distribution will change from approximate homogeneity to imbalanced distribution and finally to extreme heterogeneity.

\subsection{Modality Alignment}
\label{encoding_intro}
In multimodal learning, data from different modalities (such as vision and language) is typically aligned within a shared embedding space to enable representation learning.

\noindent \textbf{Text Encoding.}
We firstly tokenize the text description (\eg `a photo of a [classname]') and then project them into an initial set of text embeddings, denoted as $[w_0^j]_{j=1}^{N}$. These embeddings are then processed by a text encoder $\mathcal{T}$ through a series of blocks. In each block, the embeddings undergo the following transformation:
\begin{equation}
    [w_i^j]_{j=1}^{N} = \mathcal{T}_i([w_{i-1}^j]_{j=1}^{N}) \quad i \in [1, L],
\end{equation} 
where $i$ represents the block index. This process continues through the final block, resulting in the $L$-th text embedding $w_L^N$. Then we project it into a shared vision-language latent space, yielding the text feature $z_l = \mathbb{E}_T(w_L^N)$.

\noindent \textbf{Image Encoding.}
An image $x$ is first divided into $M$ fixed-size patch embeddings $E_0$, which are fed into $L$-blocks image encoder $\mathcal{V}$ alongside a learnable class token $cls_0$. At each block, the concatenated feature $[cls_{i-1}, E_{i-1}]$ is passed sequentially through the vector $\mathcal{V}_i$ as follows:
\begin{equation}
    [cls_i, E_i] = \mathcal{V}_i([cls_{i-1}, E_{i-1}]) \quad i \in [1, L].
\end{equation}
Then, a projection layer $\mathbb{E}_I$ maps the resulting class token $cls_L$ into a shared vision-language latent space to produce the image feature $z_v=\mathbb{E}_I(cls_L)$.

\noindent \textbf{Remark.}
Before classification, feature normalization is applied to both the image and text embeddings, defined as \( \bar{z}_v = \frac{z_v}{\| z_v \|_2} \) and \( \bar{z}_l = \frac{z_l}{\| z_l \|_2} \), respectively. Given the normalized image feature \( \bar{z}_v \) and text feature \( \bar{z}_l \), the prediction probability \( p_{\theta} \) for each class \( c \) associated with the image \( x \) is calculated by using the cosine similarity function \( \text{sim}(\cdot, \cdot) \):
\begin{equation}\label{eq:clip_prob}
    p_{\theta}(\hat{y} = c | x) = \frac{\exp(\text{sim}(\bar{z}_v, \bar{z}_l^c) / \tau)}{\sum_{j=1}^{C} \exp(\text{sim}(\bar{z}_v, \bar{z}_l^j) / \tau)},
\end{equation}
where \( \tau \) represents a temperature parameter and \( \theta \) denotes the model parameters. The predicted class is then identified as the one with the highest probability score.

\section{FedAlign}

In this paper, we consider a federated scenario involving a central server and $K$ clients, where each client $k$ has a private vision dataset $\mathcal{D}_k=\{(x_i, y_i)\}_{i=1}^{N_k}$ with $N_k$ samples and corresponding labels.
Data distribution across clients is usually heterogeneous, such as imbalanced distributions and disjoint classes.
The goal of our framework is to optimize personalized models that perform well on the local data distribution and simultaneously construct a robust global model across the heterogeneous data distribution of all clients.
In subsequent sections, we will first present the streamlined framework, followed by a detailed explanation of each component.

\subsection{Overview}
Our framework consists of three main components: \textit{Personalized Local Updating}, \textit{Knowledge Sharing and Global Training}, and \textit{Parameter Aggregation}. Inspired by CLIP \cite{CLIP}, we issue the image encoder to each client for performing personalized model parameter training with local private images. Later, each client uploads the image features with corresponding labels to the server. To address heterogeneous challenges in FL, our framework migrates the text encoder to the server side and performs the global text encoder training. Furthermore, clients also share the class prototypes and low-rank weight parameters for knowledge interaction and aggregation. To mitigate overfitting and maintain performance efficiency between global and personalized models, the server aggregates parameters for each client and broadcasts updates back to clients.

\subsection{Personalized Local Updating}
Each client has access to a pre-trained image encoder of the off-the-shelf CLIP, \eg ViT-B/32.
With its strong representations, clients can improve robustness and stability on heterogeneously distributed data during local training.
Recent studies \cite{PromptFL, FedOTP, FedPGP} often leverage prompt learning to fine-tune personalized models or a unified global model to reduce computation and communication burdens.
However, these methods need to select the appropriate prompt to adapt tasks during the inference stage, which increases inference latency or does not fully account for both effective global and local performances.
We notice that the LoRA mechanism introduces additive low-rank matrices to original weights, allowing the parameter addition between the model updates which correspond to downstream tasks and frozen backbone parameters. 
Inspired by it, we focus on training personalized models to adapt local data distributions and aggregating a global model through its additive parameter composability without extra inference overhead.

\noindent \textbf{Low-rank Updating.}
For any self-attention weight matrix ${W}$ in the local image encoder $\mathcal{V}_k$, the matrix consists of a pre-trained weight ${W_0} \in {\mathbb{R}}^{d_1 \times d_2}$ of $\mathcal{V}_k$ and two learnable small matrices ${A} \in {\mathbb{R}}^{r \times d_2}$ and ${B} \in {\mathbb{R}}^{d_1 \times r}$:
\begin{equation}
   {W} = {W_0} + {BA}, 
\end{equation}
where the rank $r \ll \{d_1, d_2\}$ and the client can control computational overhead by appropriately adjusting it.
Following the previous setting \cite{CLIP_LoRA}, we use Kaiming and zero initialization for $A$ and $B$, respectively.
Client $k$ then scales the matrix updates $\Delta {W_k}={B_kA_k}$ by a scaling factor $\gamma$.
During the training phase, ${A_k}$ and ${B_k}$ can receive gradient updates while keeping ${W_0}$ frozen, further reducing computational overhead.
Notably, only the self-attention modules are learnable while MLP modules are frozen in local image encoders for parameter-efficiency.

To encourage all learnable matrices $\Delta \mathbf{W_k}$ of $\mathcal{V}_{k}$ to adapt to downstream tasks, we employ the cross-entropy loss over local data samples $\mathcal{D}_k$ and the loss is calculated by:
\begin{equation}
   \mathcal{L}_{ce}= -\frac{1}{N_k} \sum_{(x,y) \in \mathcal{D}_k} \sum_{c} y \log p_{\theta_k}(\hat{y}=c|x), 
\end{equation}
where the prediction probability $p$ refers to Eq.(\ref{eq:clip_prob}) and the text feature $\bar{z}_l$ in it are retrieved from the central server.
The model parameter $\theta_{k}$, which includes the frozen backbone parameter $\mathbf{W}_0$ and learnable matrices $\Delta \mathbf{W}_k$, represents as $[\mathbf{W}_0, \Delta \mathbf{W}_k]$.
Furthermore, we want to further facilitate classification performance by scaling up the feature embedding distance between different classes.
To this end, we introduce an orthogonality penalty loss \cite{CoDA_Prompt} to local training:
\begin{equation}
   \mathcal{L}_{orth} = \| \bar{Z}_v \bar{Z}_v^T- I\|_2, 
\end{equation}
where $\bar{Z}_v$ represents a vector set of per-class normalized image features in a batch, \ie, all normalized image features $\bar{z}_v$ in the batch are averaged by category and then concatenated together.
Given local samples and fetched text features, our local optimization objective for client $k$ is:
\begin{equation}
\label{eq:local_encoder_update}
   \min_{\Delta \mathbf{W}_k} \mathcal{L}_{ce}([\mathbf{W}_0,\Delta \mathbf{W}_k]; \mathcal{D}_k) + \mu \mathcal{L}_{orth},
\end{equation}
where $\mu$ is a hyper-parameter to balance the loss $\mathcal{L}_{orth}$.

\subsection{Knowledge Sharing and Global Training}
\label{text_description}

\noindent\textbf{Knowledge Sharing.} 
After local training, clients need to upload visual features with corresponding labels for text encoder training and updated weight matrices of local image encoders for parameter aggregation.
Precisely, clients only share correctly predicted features with labels, which we denoted as $\tilde{\mathcal{D}}_k=\{(\bar{z}_{v},\hat{y})_i\}_{i=1}^{\tilde{N}_k}$ and $\tilde{N}_k$ is the number of correct predictions.
Furthermore, clients only upload updated weight parameters $\Delta \mathbf{W}_k$ instead of the entire model parameter to reduce communication costs.
Besides, clients calculate the per-class prototype $u_{k,c}$ for global and personalized image encoder parameter aggregation on the server side.
Similar to the previous definition~\cite{FedProto, FedTGP, FedPCL, FPL, FedPL}, we consider the prototype of class $c$ by averaging feature embeddings $\bar{z}_{v}$ that are predicted correctly:
\begin{equation}
\label{eq:local_prototype}
    u_{k,c} = \frac{1}{\tilde{N}_{k,c}}\sum_{(\bar{z}_{v},\hat{y}) \in \tilde{\mathcal{D}}_k, \hat{y}=c} \bar{z}_{v},
\end{equation}
where $\tilde{N}_{k,c}$ is the sample number of class $c$ in $\tilde{N}_k$. 

\noindent \textbf{Modality-driven Global Training.}
Given text descriptions $text_c$ which only relate to the category name, the text encoder $\mathcal{T}$ can generate global text features $\bar{z}_l$ refer to ``Text encoding'' in Section~\ref{encoding_intro}.
The description has multiple generation styles, we briefly consider two ways: 1) \textit{ST}: simply apply the sentence `a photo of a [classname]'; 2) \textit{GT}: exploit a pre-trained large language model LLM~\cite{GPT, CoPrompt} to generate a detailed textual description `GPT([classname])'.
Similar to local training in image encoders, the global text encoder is optimized by the LoRA mechanism with a cross-entropy loss given uploading image feature sets $\{\tilde{\mathcal{D}}_k\}_{k=1}^{K}$:
\begin{equation}
\label{eq:text_encoder_update}
    \min_{\Delta \mathbf{W}_{text}} \mathcal{L}_{ce}([\mathbf{W}_{0,text}, \Delta \mathbf{W}_{text}];\{\tilde{\mathcal{D}}_k\}_{k=1}^{K}, text_c),
\end{equation}
where $\Delta \mathbf{W}_{text}=\mathbf{B}_{text}\mathbf{A}_{text}$ is learnable text weight matrices and $\mathbf{W}_{0,text}$ is the frozen pre-trained text encoder parameters.
Using decoupling model training, the text encoder can be updated without private images from clients, reducing the overfitting effect caused by the entire CLIP model training on a single client.
After optimization, the server distributes global text features $\{\bar{z}_l^c\}_{c=1}^{C}$ to each client.

\subsection{Parameter Aggregation}
\label{parameter_aggregation}

\noindent\textbf{Query-based Aggregation.}
We propose a query-based parameter aggregation mechanism over uploading weights $\{\Delta \mathbf{W}_k\}_{k=1}^{K}$ to update different local image encoders and construct a global image encoder to maintain global performance.
Specifically, each client $k$ can learn class-related transferable knowledge from other clients, even their local samples across different categories. The key is to effectively quantify knowledge relevance between visual features from different clients. Inspired by~\cite{Transformer, EvoPrompt}, we treat class-specific prototypes $u_{k,c}$ of all classes in client $k$ as the query, and prototypes from other clients as keys. We compute the similarity between each query and keys, then we sum them to derive the relational attention factor $d_{kj}^c$ of all knowledge that can be learned from client $j$ for client $k$ at local class $c$:
\begin{equation}
    d_{kj}^c = \sum_{c'=1}^{C_j} sim(u_{k,c},u_{j,c'}),
\end{equation}
where $sim(\cdot, \cdot)$ presents the similarity function (usually cosine similarity) and $C_j$ is the class number of client $j$.
Then, the influence factor of any client $j$ to client $k$ is defined as:
\begin{equation}
    \alpha_{kj} = \sum_{c=1}^{C_k} \frac{N_{k,c}}{N_k} d_{kj}^{c},
\end{equation}
where $N_{k,c}$ is a positive constant that represents the sample number of the class $c$ on client $k$.
For each client $k$, we take the client-specific weight matrices $\Delta \mathbf{W}_j$ as the value, and the factor $\alpha_{kj}$ controls the contribution of value aggregation from different user $j$.
Concisely, the coefficient vector for client $k$ is denoted as $\alpha_k = [\alpha_{k1}, \cdots, \alpha_{kK}]$ with the sum of elements equals 1 after performing softmax operation.
Finally, the server aggregates weight matrices for each client that serve as a good initialization of every global round to facilitate local adaptation:
\begin{equation}
\label{eq:aggregate_image_encoder}
    \Delta \mathbf{W}_{k,ini}^{t+1} = \alpha_k \Delta \mathbf{W}^{t},
\end{equation}
where $\Delta \mathbf{W}^t=[\Delta \mathbf{W}_1^t, \cdots, \Delta \mathbf{W}_K^t]^T$ is the weight parameter set uploading from clients at the global round $t$ finished.

\noindent \textbf{Weighted Aggregation.}
For the global image encoder $\mathcal{V}_g$, we utilize a simple weighted aggregation mechanism to generate the global updated parameter:
\begin{equation}
\label{eq:aggregate_global_image_parameter}
    \Delta \mathbf{W}_g^{t+1} = \sum_{j=1}^K \frac{N_j}{\sum_{j=1}^K N_j} \Delta \mathbf{W}_j^t.
\end{equation}
The entire parameter of $\mathcal{V}_g$ is denoted as $[\mathbf{W}_0, \Delta \mathbf{W}_g^{t+1}]$, which the frozen backbone parameter $\mathbf{W}_0$ is invariable.

\noindent \textbf{Remark.}
Furthermore, the latent representations in an image encoder are divided by low-level and high-level space according to the embedding discrepancy of different categories.
The former is focused on general visual features while the latter tends to abstract features like category.
According to this, we only employ the query-based aggregation mechanism to generate personalized parameter updates on high-level space and apply global parameter updates on the low-level space for getting general representations.
Thus, the personalized parameter matrices are then formalized as $\Delta \mathbf{W}_{k,p}^{t+1} = [\{\Delta \mathbf{W}_{g}^{t+1}\}_{l}^{m-1}, \{\Delta \mathbf{W}_{k,ini}^{t+1}\}_{m}^{L}]$, where $l$ is the LoRA starting layer index and $m$ is the boundary layer.
To reduce computation, we can start LoRA updating not from the first layer of the image encoder but from the $l$-th layer.
After every global round, the server delivers updated personalized matrices $\Delta \mathbf{W}_{k,p}^{t+1}$ to the corresponding client $k$.

\noindent \textbf{Ex-query Aggregation.}
According to \cite{FCCL+}, clients tend to focus on local data distributions in local training, leading to the knowledge forgetting what they have learned from others. 
To address this, the server excludes the self-weight contribution during personalized aggregation $\Delta \textbf{W}_{k,ini}^{t+1}$ for client $k$ (\ie, $\alpha_{kk} = 0$), thus increasing the relative weight of external knowledge and enhancing the global performance.
Thanks to the additive ability of the LoRA mechanism, the adaptability of global and local performances can be satisfied simultaneously by adding different updated parameters (\ie $\Delta \mathbf{W}_g^{t+1}$ and $\Delta \mathbf{W}_{k,p}^{t+1}$, respectively) without inference delay, instead of seeking global and local balance under one set of updating parameters.

\subsection{Inference}
Once the $T$-round global training is finished, the server will utilize the well-trained global text encoder to generate final text features. 
At the inference stage, an image encoder $\mathcal{V}_g$ with parameters $[\mathbf{W}_0, \Delta \mathbf{W}_{g}^{T}]$ is used to evaluate the global performance.
As for personalized performances of clients, the parameter of local image encoder $\mathcal{V}_k$ consists of the frozen weight $\mathbf{W}_0$ and $\Delta \mathbf{W}_{k,p}^T$.
Notably, there is no need to exclude self-weight of $\Delta \mathbf{W}_{k,p}^T$ during the inference phase.
In Algorithm~\ref{alg:algorithm}, we summarize the training details.

\begin{algorithm}
\caption{The training procedure of FedAlign}
\label{alg:algorithm}
\KwIn{Global rounds $T$; Local epochs $T_c$; Frozen model parameters $\mathbf{W}_{0,text}$, $\mathbf{W}_0$; initial $\{\Delta \mathbf{W}_{k,p}^0\}_{k=1}^{K}$ and text features $\{\bar{z}_{l}^{c,0}\}_{c=1}^{C}$.}
\KwOut{Local parameters $\{\Delta \mathbf{W}_{k,p}^T\}_{k=1}^{K}$ and global parameters $\Delta \mathbf{W}_g^T$ and $\Delta \mathbf{W}_{text}$.}

Initialization: load frozen parameters and learnable LoRA parameters to corresponding encoders.

\For{$t=0, 1, \cdots, T-1$}{
    \For{$k={1, \cdots, K}$}{
        $\{\Delta \mathbf{W}_k^t, \{u_{k,c}\}_{c=1}^{C_k}, \tilde{\mathcal{D}}_k\}\gets$\texttt{LocalUpdate}($k$) \\
    }
    Update text encoder $\mathcal{T}$ and generate $\{\bar{z}_l^{c,t}\}_{c=1}^{C}$. \hfill{$\rhd$ in Eq. $(\ref{eq:text_encoder_update})$} \\
    Generate $\Delta \mathbf{W}_{k,p}^{t+1}$ and $\Delta \mathbf{W}_{g}^{t+1}$ for image encoders. \hfill{$\rhd$ in Eq. $(\ref{eq:aggregate_image_encoder}, \ref{eq:aggregate_global_image_parameter})$} \\
    Return $\Delta \mathbf{W}_{k,p}^{t+1}$, $\{\bar{z}_l^{c,t+1}\}_{c=1}^{C}$ to each client k.
}

\SetKwFunction{myFunction}{LocalUpdate}
\SetKwProg{myProc}{Function}{}{}
\SetFuncSty{texttt}
\myProc{\myFunction{$k$}}{
    Load the client-specific LoRA parameter $\Delta \mathbf{W}_{k,p}^{t}$ and text features $\{\bar{z}_{l}^{c,t}\}_{c=1}^{C}$. \\
    \For{$e=0, 1, \cdots, T_c-1$}{
        \ForEach{batch $b$ of $\mathcal{D}_k$}{
            Update image encoder $\mathcal{V}_k$. \hfill{$\rhd$ in Eq. $(\ref{eq:local_encoder_update})$} \\
            $\forall$ $i$ in $b$, \If{$y_{i} == \hat{y}_i$ and $e=T_c-1$}{
              $\tilde{\mathcal{D}}_k = \tilde{\mathcal{D}}_k \cup (\bar{z}_{v}, y)_i, \tilde{N}_{k,y_i}$++.  
            }
            
        }
    }
    Compute prototypes $\{u_{k,c}\}_{c=1}^{C_k}$. \hfill{$\rhd$ in Eq. $(\ref{eq:local_prototype})$} \\
    \Return{$\Delta \mathbf{W}_k^t, \{u_{k,c}\}_{c=1}^{C_k}, \tilde{\mathcal{D}}_k$}.\\
}
\end{algorithm}

\section{Experiments}
\subsection{Experimental Setup}\label{sec:setup}
\noindent \textbf{Datasets and Heterogeneous Settings.}
Following previous methods in FL, we evaluate our algorithm on six public benchmark datasets (\ie CIFAR10 \cite{CIFAR}, CIFAR100 \cite{CIFAR}, EuroSAT \cite{EuroSAT}, OxfordPet \cite{OxfordPet}, Food101 \cite{Food101}, Flower102 \cite{Flower102}) over different data distributions.
To better express three data distributions in Figure~\ref{fig:dir}, we consider three settings:
1) \textit{IID}: random assign all training samples to clients, simulating the near-uniform distribution;
2) \textit{Dir}: employ Dirichlet distribution with the parameter $\alpha$ (default 0.1) to assign samples to clients, simulating the imbalanced distribution;
3) \textit{Path}: randomly allocate disjoint class samples to clients, simulating the near one-hot partition.
See the Appendix for the detailed data partition.

\noindent \textbf{Baselines.}
We compare our FedAlign with four SOTA baselines based on CLIP:
1) FedMaPLe performs FL with the latest multi-modal prompt learning technique \cite{MaPLe};
2) PromptFL \cite{PromptFL} employs a unified prompt to fine-tune the pre-trained model;
3) FedOTP \cite{FedOTP} integrates global and local prompts to facilitate model performance;
4) FedPGP \cite{FedPGP} introduces a low-rank decomposition adaptation scheme into federated prompt learning.
We run 50 global rounds for these algorithms to get better performance and convergence.

\begin{table*}[!htb]
\centering
\begin{tabular}{cc}  
    \begin{subtable}{0.44\textwidth}
        \centering
        \begin{tabular}{l|ccc}
        \toprule
        \multicolumn{1}{c|}{Method} & IID   & Dir   & Path     \\
        \hline
        FedMaPLe                    & $96.55^{96.55}$ & $93.87^{93.86}$ & $86.22^{97.93}$    \\
        PromptFL                    & $93.91^{94.04}$ & $87.13^{90.11}$ & $81.66^{96.68}$    \\
        FedOTP                      & $92.14^{91.92}$ & $59.74^{84.48}$ & $20.00^{99.46}$    \\
        FedPGP                      & $94.42^{94.26}$ & $75.72^{91.35}$ & $19.96^{99.71}$    \\
        FedAlign                    & $\mathbf{97.66}^{\mathbf{97.66}}$ & $\mathbf{96.76}^{\mathbf{97.26}}$ & $\mathbf{94.38}^{\mathbf{99.83}}$    \\
        \bottomrule
        \end{tabular}
        \caption{CIFAR10}
    \end{subtable}
    &\quad
    \begin{subtable}{0.44\textwidth}
        \centering
        \begin{tabular}{l|ccc}
        \toprule
        \multicolumn{1}{c|}{Method} & IID   & Dir   & Path    \\
        \hline
        FedMaPLe                    & $82.64^{82.69}$ & $68.88^{75.45}$ & $66.95^{66.95}$    \\
        PromptFL                    & $73.87^{74.02}$ & $62.11^{67.14}$ & $60.57^{60.57}$    \\
        FedOTP                      & $65.48^{64.95}$ & $28.69^{52.36}$ & $17.83^{85.24}$    \\
        FedPGP                      & $75.55^{75.08}$ & $40.21^{69.49}$ & $20.25^{90.58}$    \\
        FedAlign                    & $\mathbf{85.63}^{\mathbf{85.64}}$ & $\mathbf{79.56}^{\mathbf{82.09}}$ & $\mathbf{77.20}^{\mathbf{94.20}}$    \\
        \bottomrule
        \end{tabular}
        \caption{CIFAR100}
    \end{subtable}
    \\  
    
    \begin{subtable}{0.44\textwidth}
        \centering
        \begin{tabular}{l|ccc}
        \toprule
        \multicolumn{1}{c|}{Method} & IID   & Dir   & Path     \\
        \hline
        FedMaPLe                    & $97.52^{97.57}$ & $87.85^{89.04}$ & $64.69^{84.25}$    \\
        PromptFL                    & $91.63^{91.23}$ & $74.02^{79.71}$ & $44.13^{76.69}$    \\
        FedOTP                      & $90.14^{89.82}$ & $51.74^{79.16}$ & $19.73^{98.68}$    \\
        FedPGP                      & $93.66^{92.59}$ & $55.19^{84.20}$ & $19.87^{99.18}$    \\
        FedAlign                    & $\mathbf{98.39}^{\mathbf{98.39}}$ & $\mathbf{96.98}^{\mathbf{97.30}}$ & $\mathbf{89.87}^{\mathbf{99.71}}$    \\
        \bottomrule
        \end{tabular}
        \caption{EuroSAT}
    \end{subtable}
    &\quad
    \begin{subtable}{0.44\textwidth}
        \centering
        \begin{tabular}{l|ccc}
        \toprule
        \multicolumn{1}{c|}{Method} & IID   & Dir   & Path     \\
        \hline
        FedMaPLe                    & $89.40^{89.00}$ & $87.57^{88.50}$ & $81.55^{93.40}$    \\
        PromptFL                    & $87.08^{86.99}$ & $85.57^{86.32}$ & $\mathbf{84.46}^{87.93}$    \\
        FedOTP                      & $71.24^{70.05}$ & $28.81^{58.91}$ & $17.77^{93.74}$    \\
        FedPGP                      & $88.16^{87.72}$ & $65.02^{89.50}$ & $59.38^{95.19}$    \\
        FedAlign                    & $\mathbf{92.23}^{\mathbf{92.27}}$ & $\mathbf{88.39}^{\mathbf{90.59}}$ & $82.04^{\mathbf{97.78}}$   \\
        \bottomrule
        \end{tabular}
        \caption{OxfordPet}
    \end{subtable}
    \\  
        
    \begin{subtable}{0.44\textwidth}
        \centering
        \begin{tabular}{l|ccc}
        \toprule
        \multicolumn{1}{c|}{Method} & IID   & Dir   & Path     \\
        \hline
        FedMaPLe                    & $86.67^{\mathbf{86.77}}$ & $77.32^{77.86}$ & $74.93^{87.48}$    \\
        PromptFL                    & $85.47^{85.45}$ & $79.37^{79.55}$ & $71.69^{82.39}$    \\
        FedOTP                      & $70.62^{70.06}$ & $28.79^{48.83}$ & $17.76^{90.04}$    \\
        FedPGP                      & $86.14^{85.88}$ & $49.60^{76.59}$ & $20.86^{\mathbf{95.59}}$    \\
        FedAlign                    & $\mathbf{86.74}^{86.72}$ & $\mathbf{80.08}^{\mathbf{83.15}}$ & $\mathbf{78.29}^{94.78}$    \\
        \bottomrule
        \end{tabular}
        \caption{Food101}
    \end{subtable}
    &\quad
    \begin{subtable}{0.44\textwidth}
        \centering
        \begin{tabular}{l|ccc}
        \toprule
        \multicolumn{1}{c|}{Method} & IID   & Dir   & Path     \\
        \hline
        FedMaPLe                    & $89.41^{91.02}$ & $77.06^{82.69}$ & $73.43^{83.80}$    \\
        PromptFL                    & $85.59^{85.98}$ & $68.33^{75.12}$ & $69.41^{77.00}$    \\
        FedOTP                      & $55.70^{54.63}$ & $24.98^{57.12}$ & $17.85^{91.40}$    \\
        FedPGP                      & $92.49^{91.59}$ & $57.96^{88.31}$ & $40.18^{98.20}$    \\
        FedAlign                    & $\mathbf{96.67}^{\mathbf{96.59}}$ & $\mathbf{93.92}^{\mathbf{95.10}}$ & $\mathbf{92.25}^{\mathbf{98.70}}$    \\
        \bottomrule
        \end{tabular}
        \caption{Flower102}
    \end{subtable}

\end{tabular}
\caption{Global and local performance comparison (\%) of different methods on six benchmark datasets under different data settings. The superscript values represent the average local performance among clients and the \textbf{bold} shows the best result.}
\vspace{-5mm}
\label{tab:performance_comparision}
\end{table*}

\noindent \textbf{Implementation Details.}
All methods apply the pre-trained CLIP model with ViT-B/32 \cite{ViT} architecture as the model backbone.
In our FL setting, we set the global communication round as $T=10$ with the client number $K=5$.
Each client performs local training with $E=1$ with the temperature parameter $\tau=2.66$ and uses a batch size of $bs=64$ for all datasets. 
We consider Adam optimizer with a learning rate of $\eta=1e-3$ to optimize both the global text encoder and local image encoders.
We prefer to apply $GT$ text description style and the detail format is shown in Appendix.
For the hyper-parameters in the local training and aggregation process, we analyze their specific values in Section \ref{ablation_study}.
We evaluate the global performance on the entire test dataset and assess the local performance on the subset of test data corresponding to the classes possessed by each client.
See the Appendix for more details.

\subsection{Compared to State-of-the-art Results}

\noindent\textbf{FedAlign Outperforms Existing Approaches.}
We first compare our proposed FedAlign with baselines in six common datasets (See Section~\ref{sec:setup} for more details) and then summarize the performance results in Table~\ref{tab:performance_comparision}. As can be found in the table, our performance significantly surpasses existing approaches that put all of the encoders (\eg text encoder and image encoder) on edge devices under different heterogeneous settings (\textbf{IID}, \textbf{Dir}, \textbf{Path}). For example, our FedAlign achieves excellent performance even compared to the second-best results (94.38\% \textit{vs.} 86.22\%, 77.20\% \textit{vs.} 66.95\%, 89.87\% \textit{vs.} 64.69\%, etc.) under extreme heterogeneous distribution (\textit{Path}). These results demonstrate that FedAlign can effectively facilitate valid knowledge sharing based on our modality alignment strategy and generalize well across multiple datasets with robust performance. 


\begin{table*}[!htb]
\centering
\resizebox{.95\textwidth}{!}{
\begin{tabular}{l|ccc|ccc|ccc|ccc} 
\hline
\multicolumn{1}{c|}{\multirow{2}{*}{Scheme}} & \multicolumn{3}{c|}{CIFAR10} & \multicolumn{3}{c|}{CIFAR100} & \multicolumn{3}{c|}{OxfordPet} & \multicolumn{3}{c}{Flower102}  \\ 
\cline{2-13}
\multicolumn{1}{c|}{}  & IID   & Dir   & Path    & IID   & Dir   & Path     & IID   & Dir    & Path    & IID   & Dir   & Path   \\ 
\hline
\rowcolor[rgb]{0.902,0.902,0.902} FedAlign   & $97.66$ & $96.76$  & $94.38$ & $85.63$ & $79.56$ & $77.20$ & $92.23$ & $88.39$ & $82.04$ & $96.67$ & $93.92$ & $92.25$ \\
FedAlign w/o \textit{GT}        & $97.50$ & $96.70$  & $93.84$ & $85.31$ & $77.91$ & $71.18$ & $90.79$ & $86.56$ & $81.77$ & $96.27$ & $93.14$ & $87.45$ \\
FedAlign w/o $\mathcal{L}_{orth}$ & $97.53$ & $\underline{96.97}$  & $92.97$ & $84.48$ & $78.10$ & $71.66$ & $91.44$ & $83.89$ & $69.66$ & $96.67$ & $93.92$ & $88.04$ \\
FedAlign w/o ex-query       & $97.51$ & $96.49$ & $93.15$ & $83.49$ & $76.57$ & $76.90$ & $90.71$ & $85.01$ & $74.68$ & $96.27$ & $92.35$ & $84.71$ \\
\hline
\end{tabular}
}
\caption{Global accuracy (\%) of ablation study on text description, additional loss, and weight selection for training and aggregation.}
\label{tab:ablation_study1}
\end{table*}


\begin{figure*}[!htb]
  \centering
  \begin{subfigure}{0.24\linewidth}
    \includegraphics[width=\linewidth]{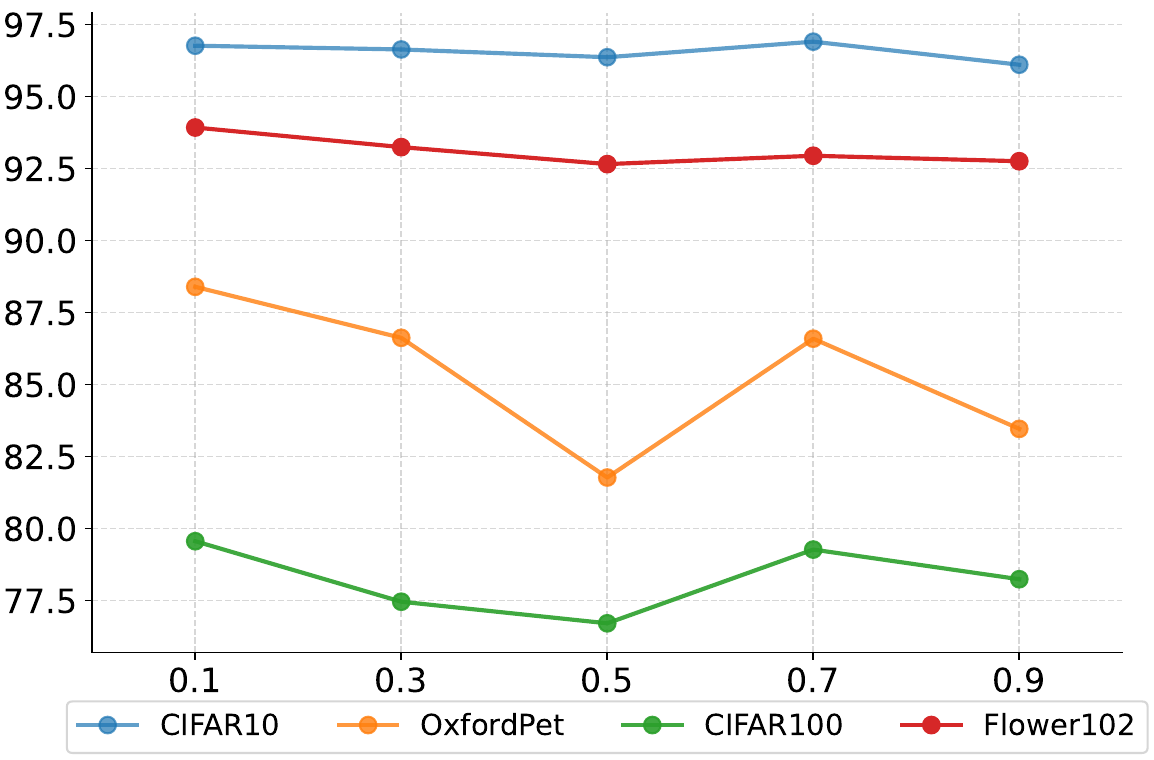}
    \caption{$\mu$}
    \label{fig:mu}
  \end{subfigure}
  \hfill
  \begin{subfigure}{0.24\linewidth}
    \includegraphics[width=\linewidth]{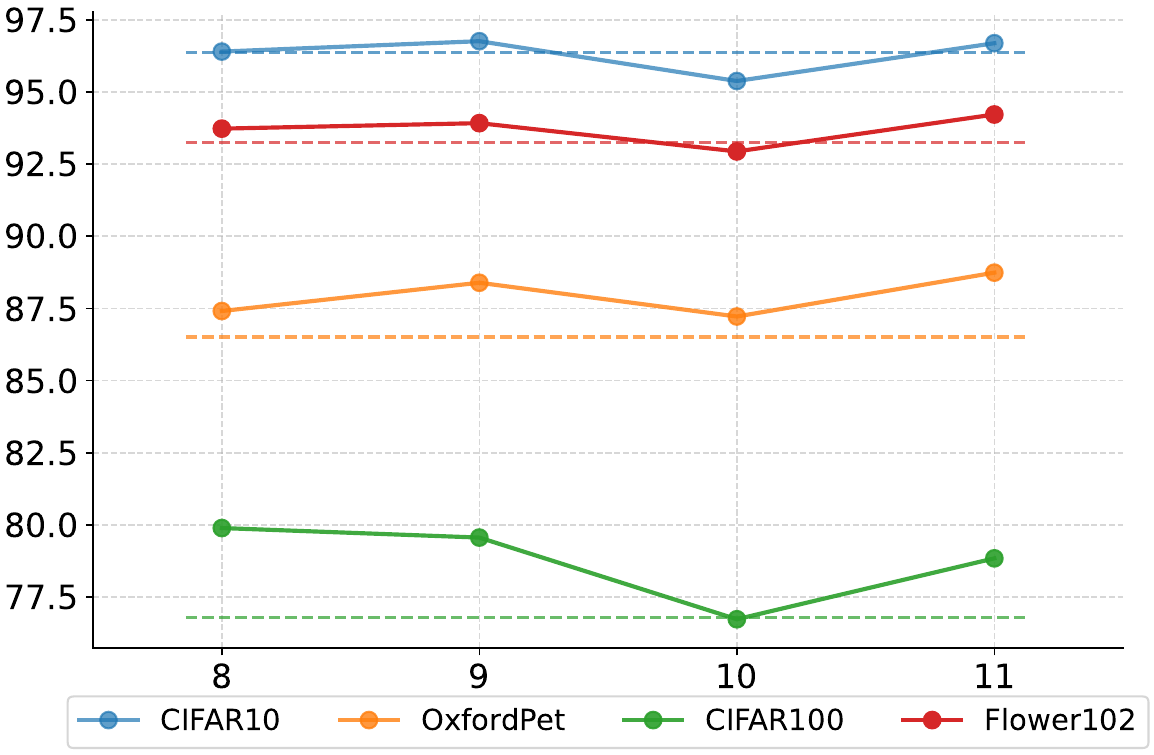}
    \caption{boundary layer}
    \label{fig:boundary_layer}
  \end{subfigure}
  \hfill
  \begin{subfigure}{0.24\linewidth}
    \includegraphics[width=\linewidth]{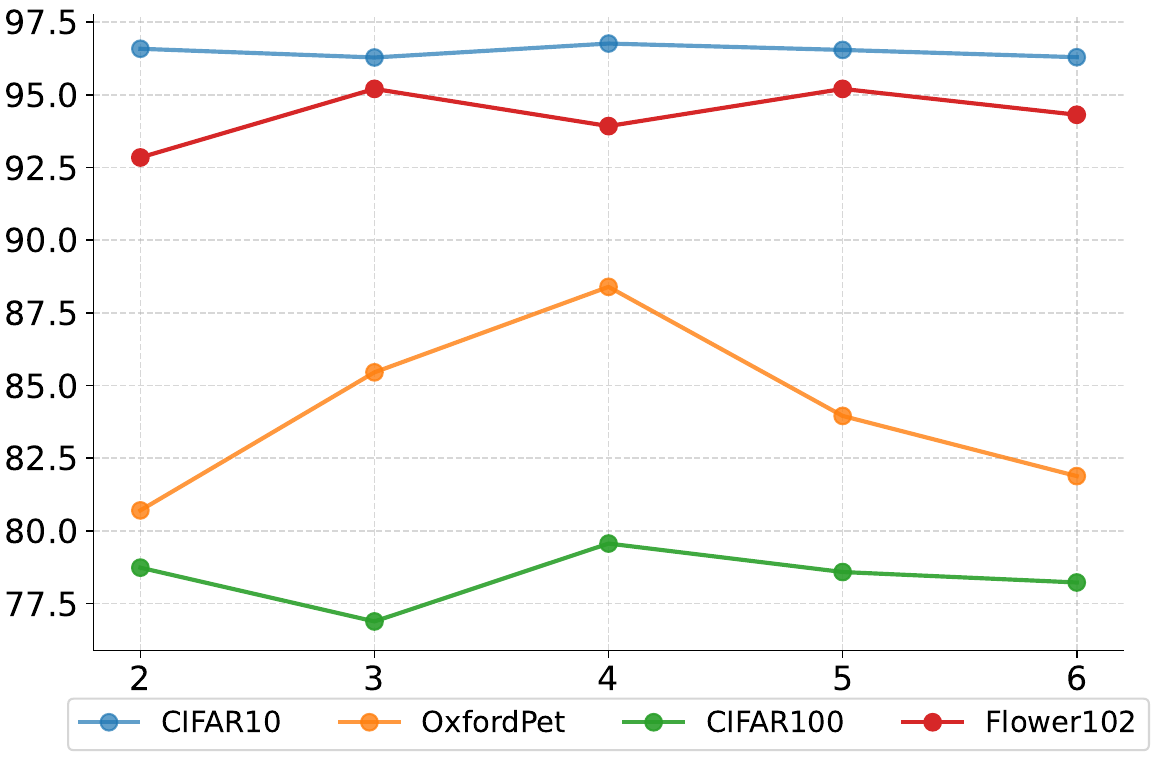}
    \caption{$r$}
    \label{fig:lora_rank}
  \end{subfigure}
  \hfill
  \begin{subfigure}{0.24\linewidth}
    \includegraphics[width=\linewidth]{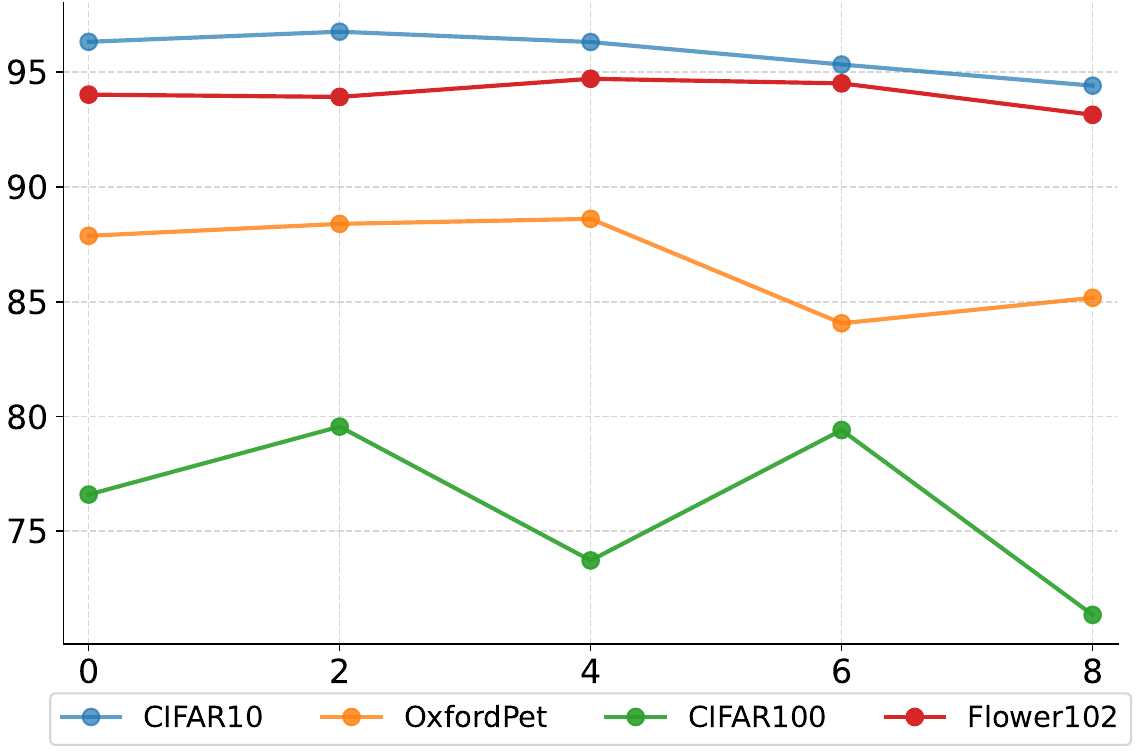}
    \caption{starting layer}
    \label{fig:balance_layer}
  \end{subfigure}
\caption{Global accuracy (\%) comparisons across different $\mu$, boundary layers, $r$, and the LoRA starting layers over the \textit{Dir} data setting. The dashed line represents our method under the pure weighted aggregation strategy.}
\vspace{-5mm}
\label{fig:ablation_theorem}
\end{figure*}

\noindent \textbf{Empowering Personalization w/o Compromising Global Capability.}
For local devices, our FedAlign also achieves robust competitive personalization results. Table~\ref{tab:performance_comparision} illustrates that there exists a significant gap between the performance of global and personalized models in existing approaches under heterogeneous data settings, particularly in the \textit{Path} setting. For instance, FedOTP and FedPGP achieve about 90\% personalized test accuracy but only reach about 20\% global accuracy on CIFAR10, CIFAR100, and EuroSAT datasets. The reason for the discrepancy may be that prompt-based algorithms emphasize adapting to local samples, leading to overfitting on local categories and hindering global performance. Apart from this, even though these approaches harness the strong representations with multi-modality strategy at edge devices, the lack of global knowledge sharing and alignment contributes to this challenge. In contrast, that is exactly the core of our approach to addressing this issue. Our approach integrates different trained parameters via the proposed aggregation mechanism with pre-trained model parameters, effectively improving global and local performance. Additional experimental results on specific details are available in the Appendix.

\subsection{Ablation Study}
\label{ablation_study}
In this section, we conduct ablation experiments on CIFAR10, CIFAR100, OxfordPet, and Flower102 datasets and provide a detailed analysis of each module and hyperparameters of our framework.
More additional results are shown in the Appendix.

\noindent \textbf{Diverse Descriptions Improve Performance.}
To investigate the impact of text encoder training, we conduct an ablation study with two kinds of text descriptions: \textit{ST} (\ie, w/o \textit{GT}) and \textit{GT} (default style) which are mentioned in Section~\ref{text_description}. 
As shown in Table~\ref{tab:ablation_study1} (\textbf{FedAlign w/o GT}, \textbf{FedAlign}), FedAlign with \textit{GT} style improves model performance due to its plentiful semantic knowledge. 
This enhancement is more pronounced in complex and heterogeneous settings, \eg, FedAlign with \textit{GT} style is 6\% more accurate than \textit{ST} style on the CIFAR100 dataset.
As a result, we adopt the \textit{GT} style to optimize text descriptions.

\noindent \textbf{Orthogonality Penalty is Effective.} 
We also evaluate the effectiveness of loss $\mathcal{L}_{orth}$ in local training by removing it and then observing the global performance fluctuation. 
As illustrated in the third row of Table~\ref{tab:ablation_study1}, the $\mathcal{L}_{orth}$ removal significantly decreases the global performance, particularly in the disjoint category setting (\ie, \textit{Path}), \eg, the performance dropped by 12.38\% and 4.21\% on OxfordPet and Flower102 datasets, respectively. 
This indicates that $\mathcal{L}_{orth}$ facilitates the separability of latent features across different classes and enhances the model classification ability.
Notably, in the \textit{Dir} setting of the CIFAR10 dataset, we observe a slight accuracy increase after removing $\mathcal{L}_{orth}$. 
We guess that it can be attributed to the simplicity and limited number of classes in this dataset, resulting in the model reaching its performance upper bound with little scope for improvement. A similar explanation applies to the unchanged accuracy can also be observed in the Flower102 dataset.

\noindent \textbf{Ex-query Promotes Aggregation.}
The fourth row of Table~\ref{tab:ablation_study1} shows global accuracy results under our FedAlign without the ex-query aggregation scheme.
Results indicate that the aggregation scheme involving their own weights participation leads to a decrease in global accuracy, with smaller declines in the \textit{IID} setting and more pronounced reductions in \textit{Dir} and \textit{Path} settings. 
This suggests that excluding the client's own weights during the aggregation process can effectively mitigate overfitting to local data distributions. 
It also improves the model's capacity to learn from other clients by reducing user-specific overfitting, as initial model parameters are less influenced by local data in the absence of the client's own weights. 

\noindent \textbf{Hyperparameters Analysis.}
\textbf{(I) Hyperparameter $\mu$ for $\mathcal{L}_{orth}$.} We conduct ablation experiments on the proportion of $\mathcal{L}_{orth}$ and results are shown in Figure~\ref{fig:mu}. We empirically set $\mu$ to 0.1 based on the validation.
\textbf{(II) Boundary Layer $m$.}
As discussed in ``Remark'' of Section \ref{parameter_aggregation}, FedAlign applies simple weighted aggregation for shallow layers which output general feature embeddings, while utilizing the query aggregation method for deeper layers which extract distinct semantic category features.
To examine the boundary layer $m$ of two aggregation mechanisms, we conduct ablation studies under $Dir$ data setting. 
Based on the results in Figure~\ref{fig:boundary_layer}, we empirically set the boundary layer to 9 due to its satisfying performance, \ie, applying query-based mechanism starting from the 9th layer, while using weighted aggregation for all preceding layers.
\textbf{(III) Rank $r$ and Starting Layer $l$.}
We explore two hyperparameters in the LoRA mechanism: the rank $r$ and the starting layer $l$ for applying LoRA. 
As shown in Figure~\ref{fig:lora_rank} and~\ref{fig:balance_layer}, we can see that:
1) The performance of CIFAR10 and Flower102 datasets is relatively robust, while the other datasets exhibit slight fluctuations;
2) The optimal hyperparameters vary across different datasets.
Considering all the observations, we select $r=4$ and the starting LoRA layer $l=2$ as a curated choice for all datasets.
\section{Conclusion}
In conclusion, we propose a federated learning framework that leverages modality alignment to bridge server-client collaboration under heterogeneous data. By decoupling a pre-trained CLIP model into a server-based text encoder and client-side image encoders, our approach supports efficient global and local optimization while preserving privacy. Using LoRA for selective parameter updates, we reduce computation and improve adaptability. An query-based aggregation mechanism enhances generalization and personalization, enabling cross-client knowledge sharing and robust performance in varied data settings. Experiments confirm that our method effectively manages data heterogeneity and outperforms traditional approaches.

{
    \small
    \bibliographystyle{ieeenat_fullname}
    \bibliography{main}
}

\setcounter{section}{0}
\renewcommand{\thesection}{\Alph{section}}
\clearpage
\setcounter{page}{1}
\maketitlesupplementary

\section{Framework of FedAlign}
\begin{figure*}[!htb]
    \centering
    \includegraphics[width=.95\linewidth]{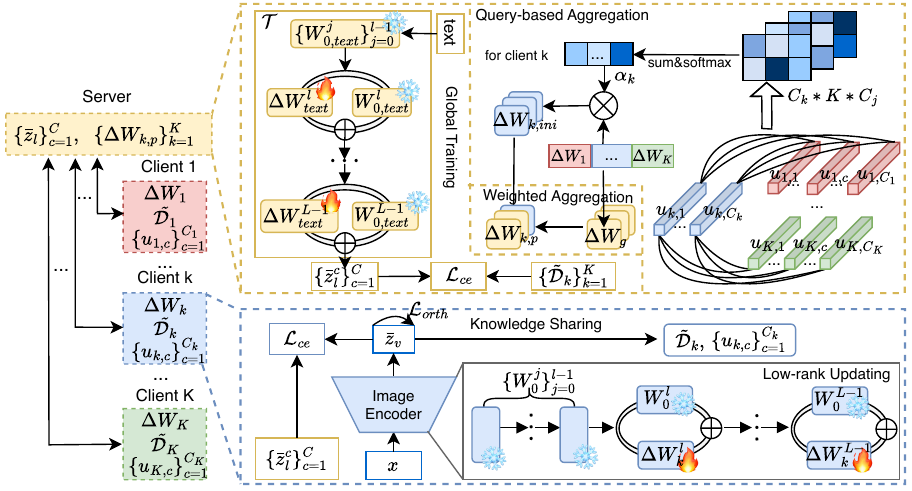}
\caption{The framework of FedAlign.}
\label{fig:framework}
\end{figure*}

In our proposed FedAlign which is shown in Figure~\ref{fig:framework}, each client optimizes the LoRA parameters $\Delta \mathbf{W}_k$ of the local image encoder $\mathcal{V}_k$ with global text features $\{\bar{z}_l\}_{c=1}^{C}$ downloaded from the central server. 
The optimization process is guided by a cross-entropy loss $\mathcal{L}_{ce}$ and an orthogonality loss $\mathcal{L}_{orth}$ for feature values with a trade-off factor $\mu$. 
To reduce computational overhead, we freeze the pre-trained backbone parameters $\mathbf{W}_0$ and set the rank of LoRA $r=4$, which is significantly smaller than the latent model dimension.
After the local training, clients complete the process of knowledge sharing by uploading three parameters to the server: the pre-class prototype $\{u_{k,c}\}_{c=1}^{C_k}$, the synthesized dataset $\mathcal{\tilde{D}}_k$ (\ie, correctly predicted image features and corresponding labels), and the updated model parameters $\Delta \mathbf{W}_k$. 

Based on uploading information, the server fine-tunes the global text encoder $\mathcal{T}$ by using the query-based aggregation mechanism.
It generates client-specific LoRA parameters $\Delta \mathbf{W}_{k,ini}$ for each client.
The server also integrates global LoRA parameters $\Delta \mathbf{W}_g$ by a weighted aggregation scheme.
To effectively utilize the feature properties of the low-level and high-level in the encoder (the low-level layers output generalized features, while the high-level model captures category features), we consider that the global parameters are used to update low-level layers, while the client-specific parameters are designated for high-level layers.
These updated parameters $\Delta \mathbf{W}_{k,p}$ are then broadcast back to the corresponding client $k$ as the initialization parameter for the local image encoder in the next training round. 
This federated process continues until convergence.
Furthermore, the ex-query aggregation and some treatments in the inference stage can enhance model performance, which we have shown in the main text.

\section{Experimental Setup}
\label{appendix_setting_details}
\subsection{Details of Computational Resources}
\label{appendix_com_resource}
Our FedAlign is implemented in PyTorch 2.4.1~\cite{PyTorch} and NVIDIA GeForce RTX 4090 with CUDA version 12.4. 

\subsection{Details of Dataset and Model Settings}
\label{appendix_model_dataset}
\noindent \textbf{Dataset Setting.}
CIFAR10~\cite{CIFAR} consists of 10 classes, each of which contains 5,000 training images and 1,000 testing images.
CIFAR100~\cite{CIFAR} contains 100 classes, with 500 training images and 100 testing images per class.
EuroSAT~\cite{EuroSAT} consists of satellite images, including 10 different classes representing various land use and land cover types with each class having 2,700 images.
OxfordPet~\cite{OxfordPet} contains 37 different breeds of cats and dogs with a total of 7,349 images. The dataset is split into a training set (about 3,680 images) and a test set (about 3,669 images).
Food101~\cite{Food101} is a dataset containing 101 classes of food images. Each class has 1,000 images, where 750 images are used for training and 250 images are used for testing.
Flower102~\cite{Flower102} contains 102 different categories of flowers. It has 8,189 images in total. The dataset is split into three subsets: 6,149 for training, 1,020 for validation, and 1,020 for testing.
The specific details of the dataset partition are shown in Table~\ref{tab:dataset_partition}.

\begin{table}[!htb]
\centering
\resizebox{0.32\textwidth}{!}{
    \begin{tabular}{l|ccc|c} 
    \hline
    Dataset        & Class   & Train  & Test   & \textit{Path} \\
    \hline
    CIFAR10        & 10      & 50000  & 10000  & 2 \\
    CIFAR100       & 100     & 50000  & 10000  & 20 \\
    EuroSAT        & 10      & 21600  & 5400   & 2 \\
    OxfordPet      & 37      & 3680   & 3669   & 7 \\
    Food101        & 101     & 75750  & 25250  & 20 \\
    Flower102      & 102     & 7169   & 1020   & 20 \\
    \hline
    \end{tabular}
}
\caption{Statistical details of benchmark datasets in experiments. We use 20\% of the EuroSAT dataset as test samples and combine the training set with the validation set of the Flower102 dataset as local datasets for clients. Other benchmarks are divided into training and testing sets according to the original data partition. We also show the category quantity of each client in the \textit{Path} setting.}
\label{tab:dataset_partition}
\end{table}

\begin{table}[!htb]
\centering
\resizebox{0.32\textwidth}{!}{
\begin{tabular}{l|cc||l|cc} 
\hline
$r$ & Text  & Image & $l$ & Text  & Image  \\ 
\hline
2 & 0.191 & 0.500 & 0 & 0.369 & 0.623  \\
3 & 0.268 & 0.544 & 2 & 0.345 & 0.588  \\
4 & 0.345 & 0.588 & 4 & 0.322 & 0.553  \\
5 & 0.422 & 0.632 & 6 & 0.298 & 0.518  \\
6 & 0.499 & 0.676 & 8 & 0.275 & 0.483  \\
\hline
\end{tabular}
}
\caption{Trainable parameters (M) of the text and image encoder with different LoRA rank $r$ and the LoRA starting layer $l$. The default values of $r$ and $l$ are 4 and 2, respectively.}
\label{tab:trainable_parameter}
\end{table}

\begin{figure*}[!htb]
    \centering
    \begin{subfigure}{0.75\linewidth}
    \includegraphics[width=\linewidth]{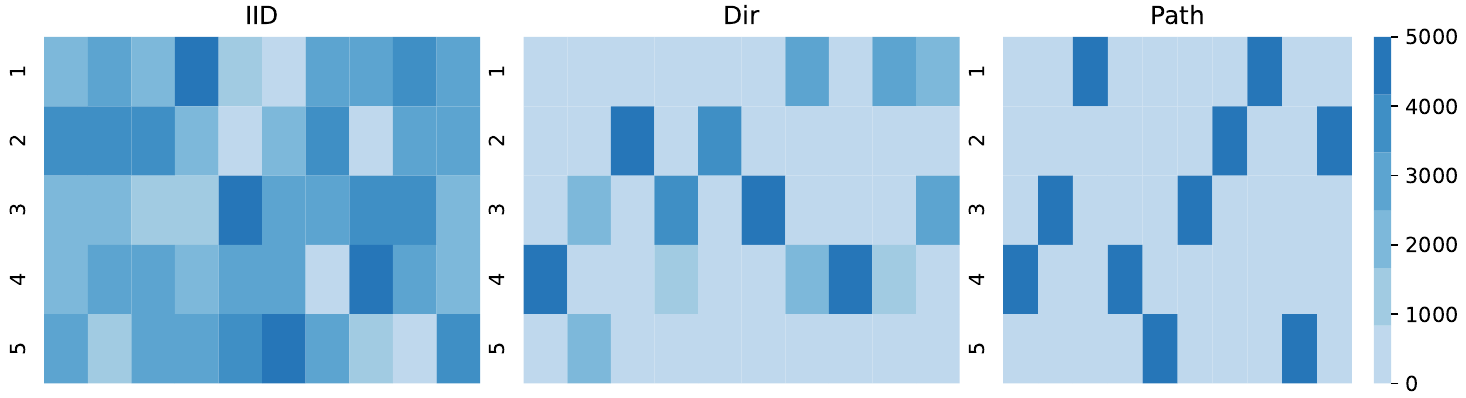}
    \caption{CIFAR10}
    \end{subfigure}
    \\
    \begin{subfigure}{0.75\linewidth}
    \includegraphics[width=\linewidth]{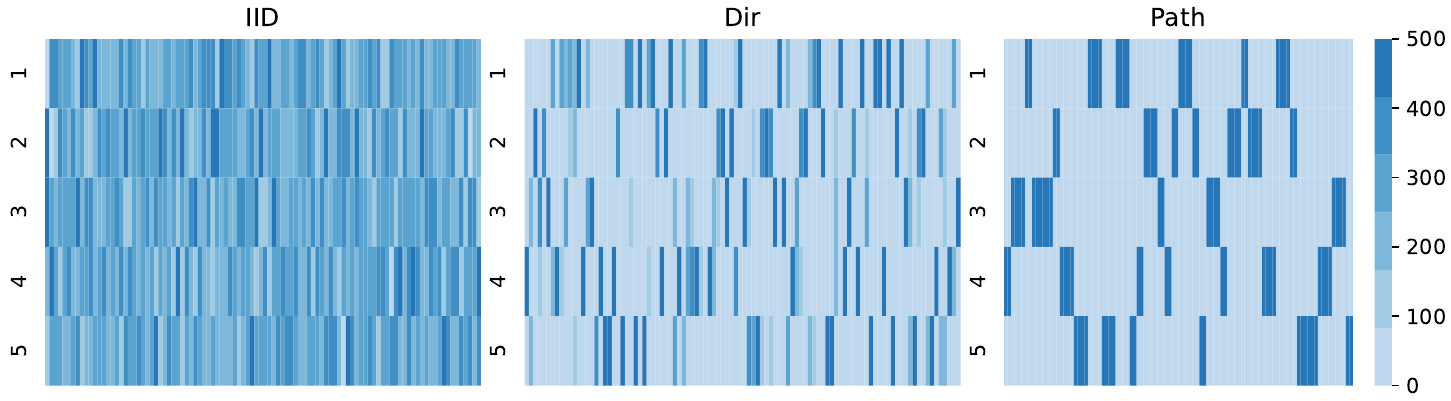}
    \caption{CIFAR100}
    \end{subfigure}
\caption{The data partition among clients. The x-axis and y-axis represent the index of the category and client, respectively.}
\label{fig:data_heatmap}
\end{figure*}

For \textbf{\textit{IID}} data setting, we randomly sample independent data from the entire training dataset.
For \textbf{\textit{Dir}} data setting, each client $k$ samples $q_{k,c} \sim Dir(\alpha)$ for each class $c$ (${\textstyle \sum_{k=1}^{K}q_{k,c}}=1$ in all datasets and $\alpha$ is a positive constant), then randomly assigns $q_{k,c}$ non-overlap proportion of samples from the benchmark training dataset for each class $c$.
In other words, for each class $c$, the number of samples in class $c$ on client $k$ (\ie, $|\mathcal{D}_{k,c}|$) is equal to $q_{k,c}*|\mathcal{D}_c|$.
In this setting, samples between clients are independent.
A higher $\alpha$ means a more balanced data distribution.
For \textbf{\textit{Path}} data setting, each client $k$ chooses disjoint class samples for training (\ie, the categories are independent between users, with no overlapping sample labels). 
We partition the entire training dataset at the category level, where each unique sample class represents a distinct subset of samples. 
Each client selects a portion of the dataset based on specific labels, ensuring that there are no overlapping labels among different clients. 
In this way, each client exclusively owns a disjoint label set from the training dataset.
\textbf{Notably, the \textit{Dir} data setting generally refers to the setting with $\alpha = 0.1$, unless otherwise specified.}
To save space, we only illustrate the data distribution of each client for the CIFAR10 and CIFAR100 datasets with different dataset settings in Figure~\ref{fig:data_heatmap}.
The data splits for the other four datasets are available in the code files.

\noindent \textbf{Model Setting.}
In our experiments, the frozen CLIP with its pre-trained parameters is based on the ViT-B/32~\cite{ViT} Transformer architecture as an image encoder and uses a masked self-attention Transformer as a text encoder. 
These encoders are trained by maximizing the similarity of (image, text) pairs through contrastive loss on large web datasets.

\begin{table*}[!htb]
\centering
\resizebox{.98\textwidth}{!}{
    \begin{tabular}{l|ccccc||ccccc}
    \toprule
    \multicolumn{1}{c|}{\multirow{2}{*}{Method}} & \multicolumn{5}{c||}{Dir(0.3)} & \multicolumn{5}{c}{Dir(0.5)} \\
    \cline{2-11}
    \multicolumn{1}{c|}{} & CIFAR10   & CIFAR100   & EuroSAT  & OxfordPet & Flower102 & CIFAR10   & CIFAR100   & EuroSAT  & OxfordPet & Flower102    \\
    \hline
    FedMaPLe   & $94.28^{95.05}$ & $76.94^{77.29}$ & $95.13^{95.66}$ & $89.81^{88.78}$ & $84.71^{87.39}$  & $93.36^{94.28}$ & $78.71^{78.19}$ & $95.35^{95.88}$ & $88.28^{88.73}$ & $88.04^{88.33}$ \\
    PromptFL   & $91.54^{92.07}$ & $66.24^{67.21}$ & $74.76^{80.40}$ & $86.05^{86.45}$ & $66.37^{67.17}$  & $90.09^{91.27}$ & $68.92^{69.29}$ & $85.35^{86.81}$ & $86.29^{86.75}$ & $78.24^{79.39}$ \\
    FedOTP     & $77.33^{80.60}$ & $41.22^{45.05}$ & $71.71^{74.61}$ & $36.64^{47.30}$ & $34.71^{46.66}$  & $84.36^{85.16}$ & $47.35^{47.67}$ & $80.39^{82.43}$ & $42.74^{46.06}$ & $37.82^{42.98}$  \\
    FedPGP     & $90.09^{89.74}$ & $60.56^{65.46}$ & $80.31^{83.99}$ & $79.54^{84.50}$ & $74.96^{87.97}$  & $90.74^{92.09}$ & $65.59^{67.06}$ & $86.86^{89.13}$ & $82.52^{83.62}$ & $85.39^{88.35}$  \\
    FedAlign   & $\mathbf{97.10}^{\mathbf{97.06}}$ & $\mathbf{82.45}^{\mathbf{82.46}}$ & $\mathbf{97.48}^{\mathbf{97.58}}$ & $\mathbf{91.28}^{\mathbf{91.22}}$ & $\mathbf{94.12}^{\mathbf{95.15}}$ & $\mathbf{97.29}^{\mathbf{97.27}}$ & $\mathbf{82.03}^{\mathbf{82.31}}$ & $\mathbf{98.09}^{\mathbf{98.09}}$ & $\mathbf{90.13}^{\mathbf{90.15}}$ & $\mathbf{95.98}^{\mathbf{96.66}}$ \\
    \hline
    \multicolumn{1}{c|}{\multirow{2}{*}{Method}} & \multicolumn{5}{c||}{Dir(0.7)} & \multicolumn{5}{c}{Dir(0.9)} \\
    \cline{2-11}
    \multicolumn{1}{c|}{} & CIFAR10   & CIFAR100   & EuroSAT  & OxfordPet & Flower102 & CIFAR10   & CIFAR100   & EuroSAT  & OxfordPet & Flower102    \\
    \hline
    FedMaPLe  & $95.72^{95.53}$ & $79.20^{79.97}$ & $96.20^{95.63}$ & $\mathbf{88.99}^{88.25}$ & $88.33^{89.52}$  & $94.89^{95.62}$ & $80.33^{80.60}$ & $97.04^{96.55}$ & $87.44^{87.95}$ & $87.25^{89.99}$ \\
    PromptFL  & $93.76^{92.41}$ & $69.80^{71.10}$ & $86.46^{86.24}$ & $86.81^{86.73}$ & $75.78^{76.48}$ & $92.63^{92.50}$ & $70.41^{71.74}$ & $88.70^{88.93}$ & $84.82^{84.43}$ & $80.20^{82.49}$ \\
    FedOTP    & $86.77^{85.66}$ & $52.04^{51.47}$ & $82.22^{81.16}$ & $49.30^{51.09}$ & $38.16^{39.66}$ & $88.03^{87.55}$ & $52.23^{51.11}$ & $83.77^{83.30}$ & $50.29^{51.38}$ & $39.43^{39.55}$ \\
    FedPGP    & $92.00^{91.25}$ & $68.40^{69.14}$ & $89.30^{87.24}$ & $85.69^{85.43}$ & $88.49^{89.09}$ & $92.83^{92.50}$ & $69.78^{70.06}$ & $91.35^{90.43}$ & $86.96^{86.54}$ & $87.90^{88.30}$ \\
    FedAlign  & $\mathbf{97.34}^{\mathbf{97.34}}$ & $\mathbf{80.24}^{\mathbf{80.41}}$ & $\mathbf{97.93}^{\mathbf{97.91}}$ & $88.80^{\mathbf{88.89}}$ & $\mathbf{96.47}^{\mathbf{96.62}}$  & $\mathbf{97.41}^{\mathbf{97.41}}$ & $\mathbf{82.99}^{\mathbf{83.03}}$ & $\mathbf{98.19}^{\mathbf{98.15}}$ & $\mathbf{90.41}^{\mathbf{90.26}}$ & $\mathbf{94.90}^{\mathbf{95.19}}$  \\
    \bottomrule
    \end{tabular}
}
\caption{Global and local performance comparison (\%) of different methods on five benchmark datasets under different data settings. The superscript values represent the average local performance among clients and the \textbf{bold} shows the best result.}
\label{tab:performance_comparision}
\end{table*}

\begin{figure*}[!htb]
  \centering
  \begin{subfigure}{0.24\linewidth}
    \includegraphics[width=\linewidth]{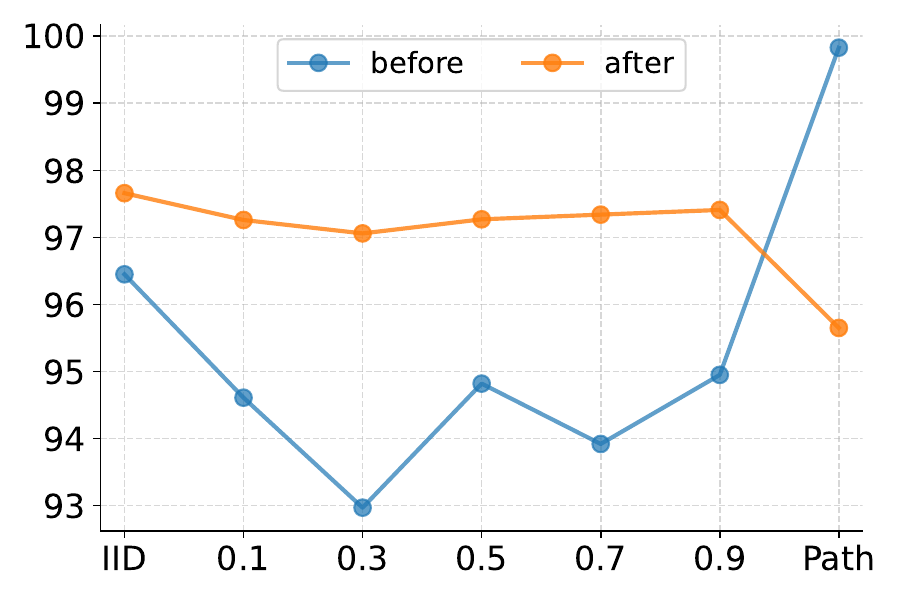}
    \caption{CIFAR10}
  \end{subfigure}
  \hfill
  \begin{subfigure}{0.24\linewidth}
    \includegraphics[width=\linewidth]{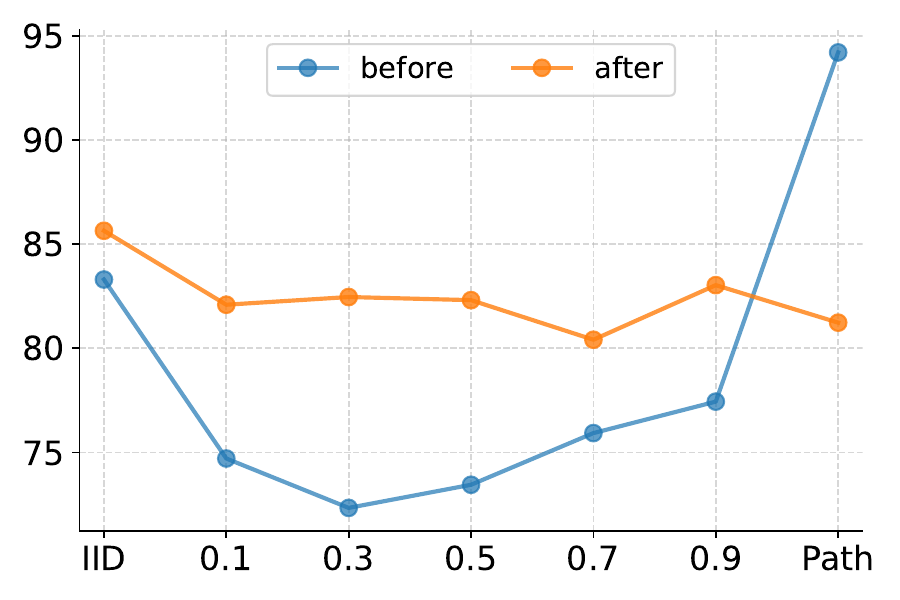}
    \caption{CIFAR100}
  \end{subfigure}
  \hfill
  \begin{subfigure}{0.24\linewidth}
    \includegraphics[width=\linewidth]{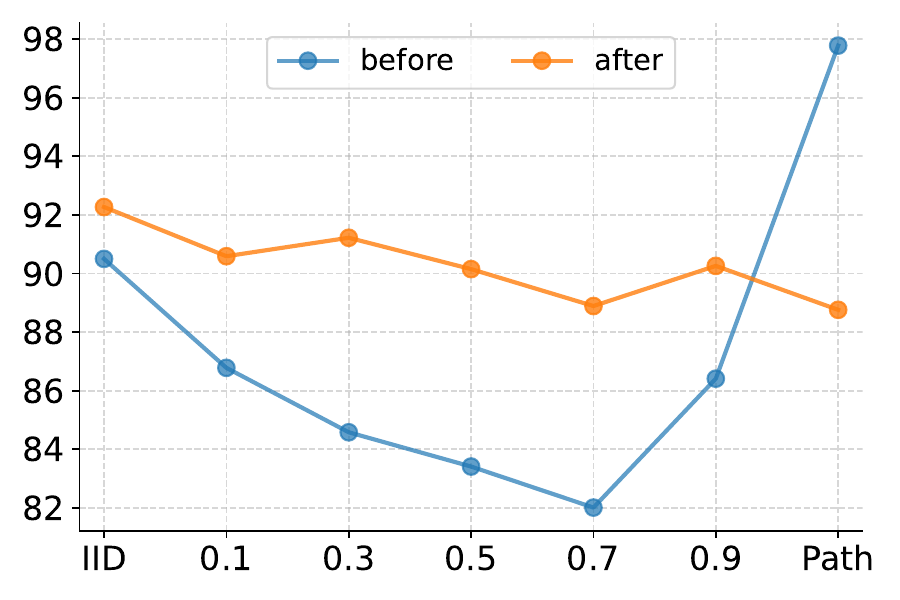}
    \caption{OxfordPet}
  \end{subfigure}
  \hfill
  \begin{subfigure}{0.24\linewidth}
    \includegraphics[width=\linewidth]{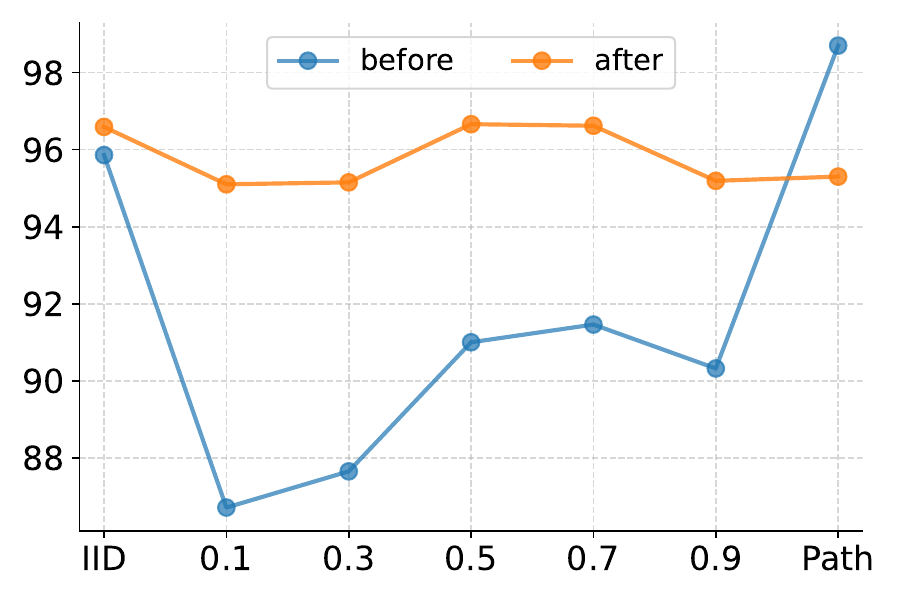}
    \caption{Flower102}
  \end{subfigure}
\caption{Average local test accuracy (\%) across different benchmark datasets over \textit{IID}, \textit{Dir}, and \textit{Path} data setting. The number on the horizontal coordinate represents the $\alpha$ in the \textit{Dir} data setting. The dashed line represents our method under the pure weighted aggregation strategy.}
\label{fig:local_aggregation_performance}
\end{figure*}

\subsection{Implementation Details}
In FedAlign, the default module uses the \textbf{\textit{GT}} style for the text description and the ex-query aggregation scheme for parameter aggregation on the server side.
The scaling factor $\gamma$ in the LoRA-based local training is a constant (\ie, 0.25).

\noindent \textbf{Dataset in the Inference Stage.}
For the global performance, we evaluate the trained global model on the entire test dataset.
For the local performance, we construct a local test set for each client by selecting the samples from the entire test data set that match the labels available in the local train dataset. 
For instance, if client $k$ only has label $1$ and label $3$ over the CIFAR10 dataset, the local testset for client $k$ will consist of all test samples with label $1$ and label $3$ over the entire CIFAR10 test dataset.
For baselines and FedAlign under the \textit{Path} setting, we utilize the local parameters obtained from the final global round to measure the local performance. 
For FedAlign under \textit{IID} and \textit{Dir} settings, we employ the parameters after the query-based aggregation to test local performance. 
The reason is that: in the \textit{IID} and \textit{Dir} settings, there are overlapping sample labels among users, so the aggregated model parameters can capture knowledge from others, thereby enhancing the local performance. 
In contrast, the \textit{Path} setting does not have overlapping sample labels, so local parameters are more appropriate for measuring local performance.
For a fair comparison, the settings of client models and the split of private datasets in all approaches are kept the same.

\subsection{Details of GT Style}
We apply a Large Language Model~\cite{GPT} to generate a short text description for each category name of all datasets. 
For example, for the label name ``bird'' in the CIFAR10 dataset, the text description in \textit{ST} style is ``a photo of a bird.'', while the description in \textit{GT} style is ``A photo of a bird, a type of animal. Birds are characterized by their feathers and beaks.''. 
For each category name, we generate multiple descriptions and randomly select one description as the input of the text encoder to produce text features. 
Detailed text descriptions for each category can be found in the code file.

\section{Experimental Results}
\subsection{Parameter Analysis}
\label{appendix_parameter_analysis}

\noindent \textbf{Trainable Parameters.}
In FedAlign, three main factors influence the number of trainable parameters: 1) the Transformer architecture of CLIP (\eg, ViT-B/32), the LoRA rank $r$, and the LoRA starting layer $l$. 
The choice of Transformer architecture determines the feature dimension in the latent space, thereby affecting the amount of trainable parameters. 
Table~\ref{tab:trainable_parameter} shows the trainable parameter count of a text and image encoder under different $r$ and $l$ in the ViT-B/32 Transformer architecture (latent dimension: 512). 
When $r$ increases or $l$ decreases, the trainable parameters in both encoders significantly rise. 
We also observe that: changes in $r$ have a more substantial impact on the trainable parameter count compared to changes in $l$. 
Furthermore, the parameter count is not affected by the number of categories, as the classification head is not involved in training.


\begin{figure*}[!htb]
    \centering
    \begin{subfigure}{0.24\linewidth}
    \includegraphics[width=\linewidth]{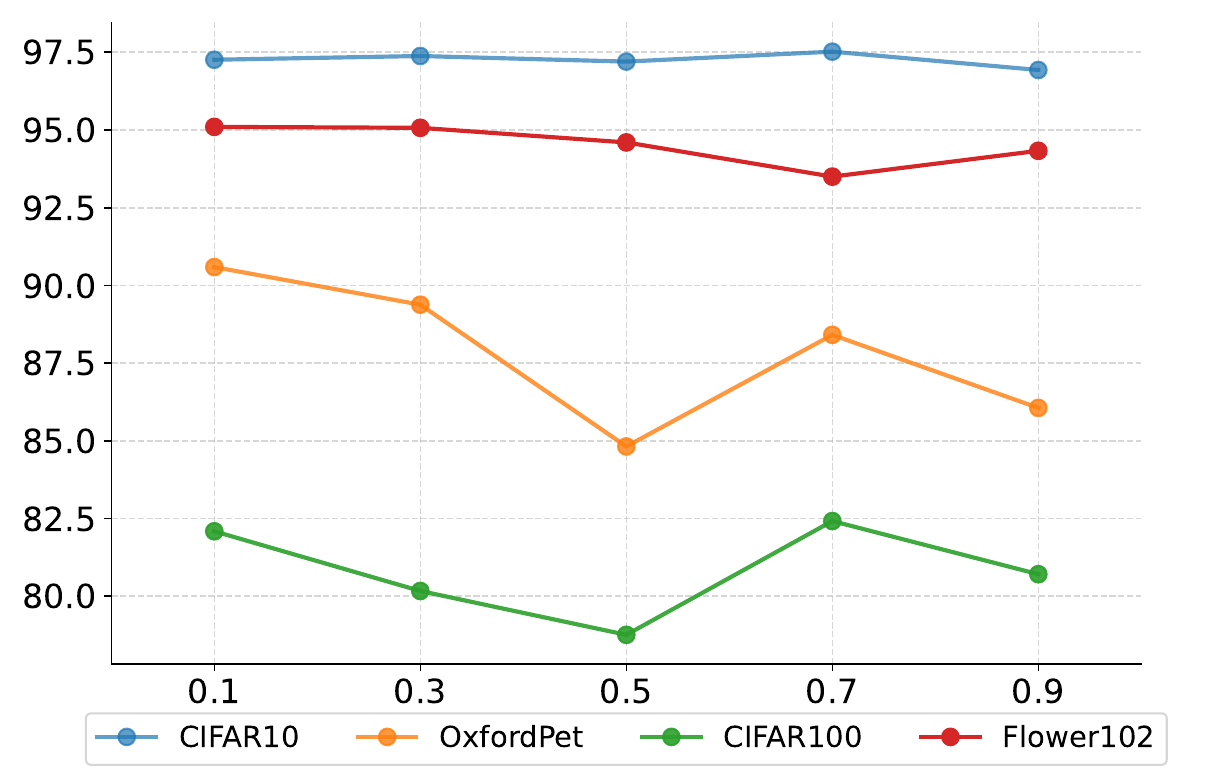}
    \caption{$\mu$}
    \label{fig:mu_local}
    \end{subfigure}
    \hfill
    \begin{subfigure}{0.24\linewidth}
    \includegraphics[width=\linewidth]{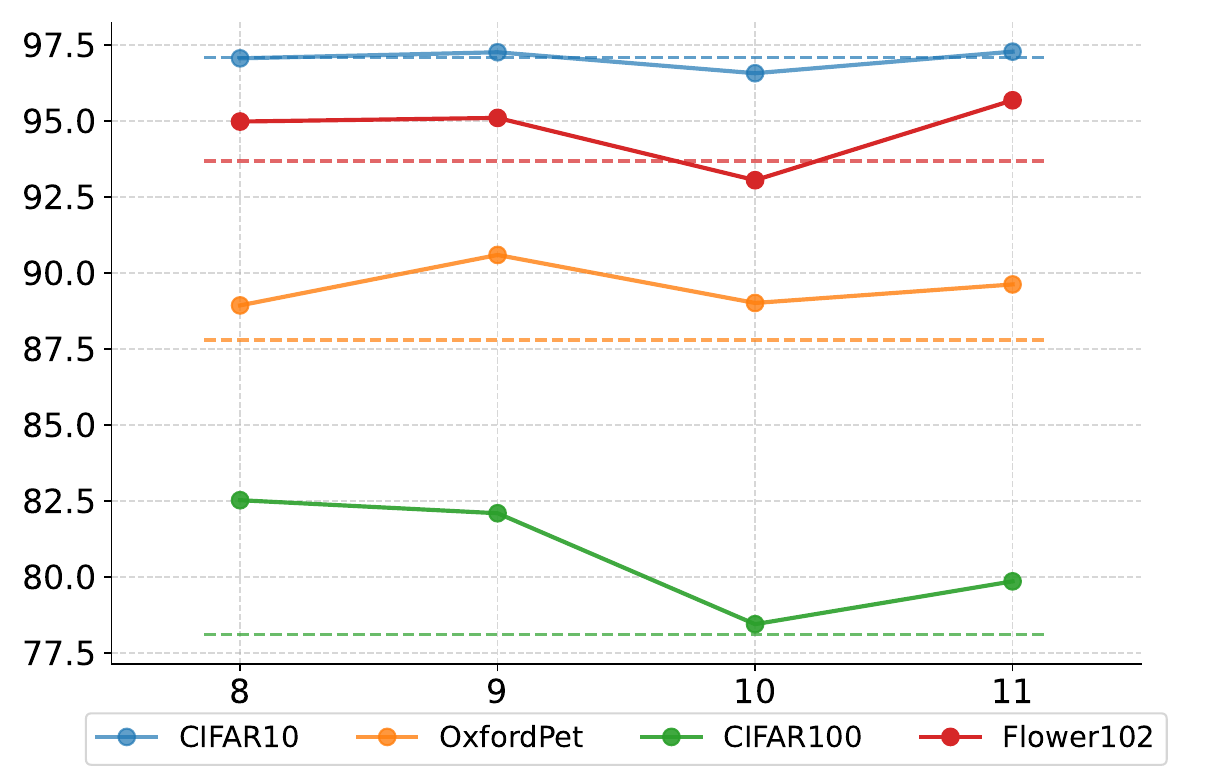}
    \caption{boundary layer}
    \label{fig:boundary_layer_local}
    \end{subfigure}
    \hfill
    \begin{subfigure}{0.24\linewidth}
    \includegraphics[width=\linewidth]{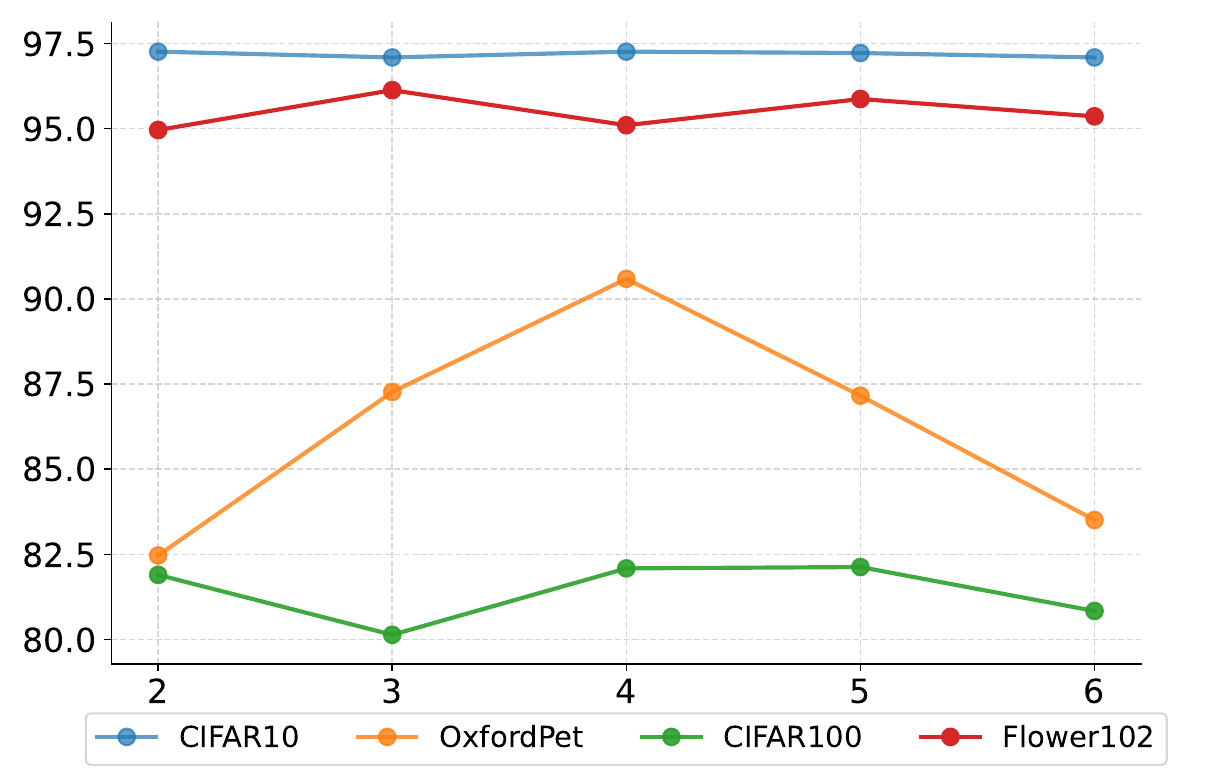}
    \caption{$r$}
    \label{fig:rank_r_local}
    \end{subfigure} 
    \hfill
    \begin{subfigure}{0.24\linewidth}
    \includegraphics[width=\linewidth]{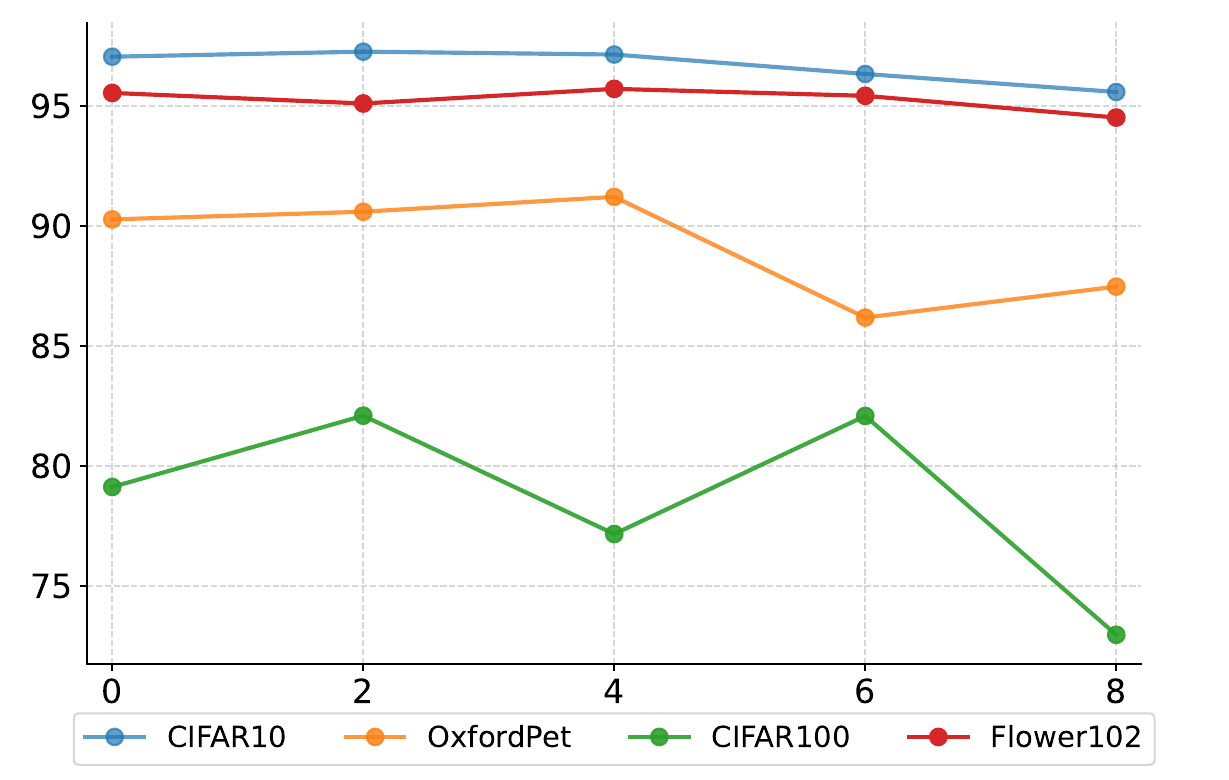}
    \caption{starting layer}
    \label{fig:starting_layer_local}
    \end{subfigure}
    \\
\caption{Average local test accuracy (\%) across different $\mu$, boundary layers, $r$, and the LoRA starting layers over the \textit{Dir} data setting. The dashed line represents our method under the pure weighted aggregation strategy.}
\label{fig:hyper_parameter_compare_local}
\end{figure*}

\begin{table}[!htb]
\centering
\resizebox{.35\textwidth}{!}{
    \begin{tabular}{l|cc|cc}
    \toprule
    \multicolumn{1}{c|}{\multirow{2}{*}{Dataset}} & \multicolumn{2}{c|}{ViT-B/32} & \multicolumn{2}{c}{ViT-B/16} \\
    \cline{2-5}
    \multicolumn{1}{c|}{} & ST   & GT   & ST  & GT  \\
    \hline
    CIFAR10   & $88.32$ & $87.85$ & $90.10$ & $88.81$ \\
    CIFAR100   & $64.47$ & $62.79$ & $68.39$ & $64.88$  \\
    EuroSAT   & $28.65$ & $44.07$ & $37.41$ & $32.20$ \\
    OxfordPet   & $80.59$ & $80.98$ & $84.85$ & $82.75$  \\
    Flower102   & $64.22$ & $60.69$ & $64.61$ & $63.82$  \\
    \bottomrule
    \end{tabular}
}
\caption{Test accuracy (\%) over the zero-shot CLIP with different Transformer architectures on five benchmark datasets. `ST' and `GT' represent the input style of the text description.}
\label{tab:zero_shot_CLIP}
\end{table}

\noindent \textbf{Communication Overhead.} 
After local training, each client needs to upload the following parameters: 1) the updated LoRA model parameters; 2) per-class prototypes, and 3) a set of image features with corresponding labels. 
The updated model parameter is the trainable parameters in the local image encoder, its parameter count is denoted as $N_{v,k}$.
The per-class prototype is related to the latent dimension (\ie, 512) and the number of local labels $C_k$.
The third component depends on the number of uploading image features that correctly predicted $\tilde{N}_{k}$ and the latent dimension of the image encoder (\ie, 512). 
Thus, the uploading parameter count in FedAlign is:
$N_{v,k}+512C_k+(512+1)\tilde{N}_{k}$.
For example, if client $k$ uploads all image features with all categories (\ie, 100) to the server over the CIFAR100 dataset, the count is $5.529$M.
While in the default hyperparameter (\ie, $r=4, l=2$), the parameter counts of the entire text and image encoder are $60.9927$M and $84.2483$M, respectively.
Compared to sharing full model parameters, the communication overhead in FedAlign is relatively tolerable.
We investigate the effect of the amount proportion of uploaded image features on the model performance, and the detail can refer to ``Impact of the Number of Uploaded Image Features.'' in the Section~\ref{appendix_analysis_details}. 

Similarly, the server broadcasts updated parameters, consisting of the global parameter $\Delta \mathbf{W}_g$ and the client-specific parameter $\Delta \mathbf{W}_{k,ini}$, to each client $k$.
To reduce communication overhead, the server can broadcast the integration version of $\{\{\Delta \mathbf{W}_g\}_{l}^{m-1}, \{\Delta \mathbf{W}_{k,ini}\}_{m}^{L}\}$ (\ie, $\Delta \mathbf{W}_{k,p}$), where $m$ and $l$ are the boundary layer and the LoRA starting layer, respectively.
This parameter number will be a constant when the hyperparameter details of models are determined.
In addition, the server also broadcasts the global text features after training the global text encoder.
The size of text features is related to the category number and latent dimension (512 in ViT-B/32).

\subsection{Details of Model Comparison with Baselines}
\label{Appendix_table_details}

\noindent \textbf{More Results in the Dir Setting.}
We conduct additional comparison experiments between FedAlign and baselines under the \textit{Dir} data setting. 
Specifically, we vary $\alpha$ across \{0.3, 0.5, 0.7, 0.9\} ($\alpha=0.1$ in the main text). 
As shown in Table~\ref{tab:performance_comparision}, our FedAlign still achieves superior global and local performance, demonstrating robustness to data heterogeneity.

\begin{figure*}[!htb]
    \centering
    \begin{subfigure}{0.28\linewidth}
    \includegraphics[width=\linewidth]{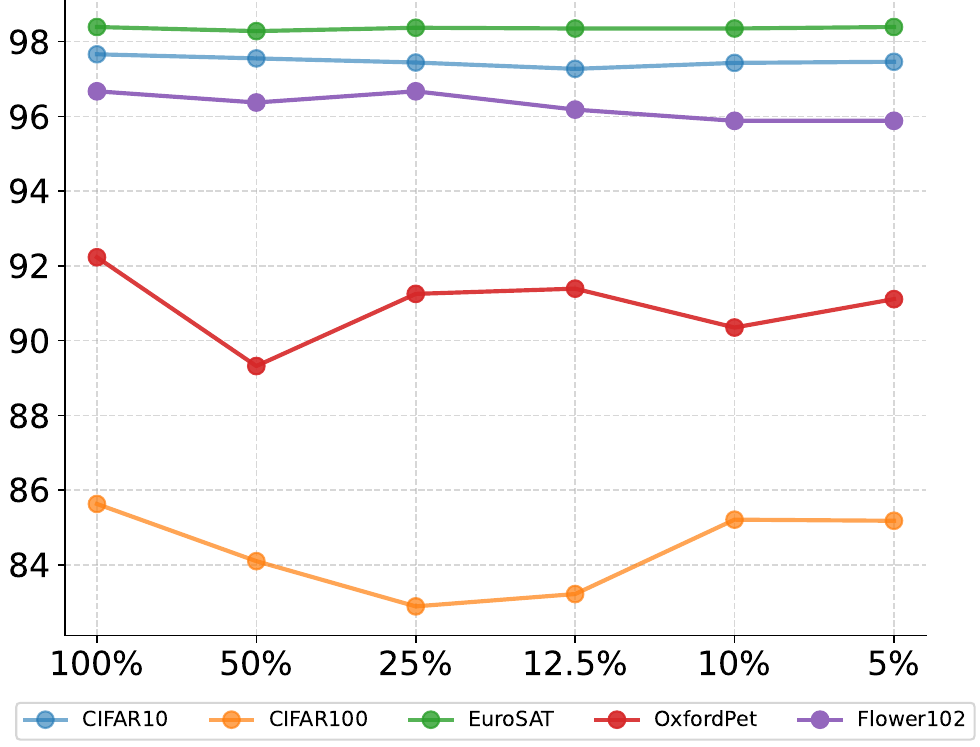}
    \caption{IID-Global}
    \end{subfigure}
    \hfill
    \begin{subfigure}{0.28\linewidth}
    \includegraphics[width=\linewidth]{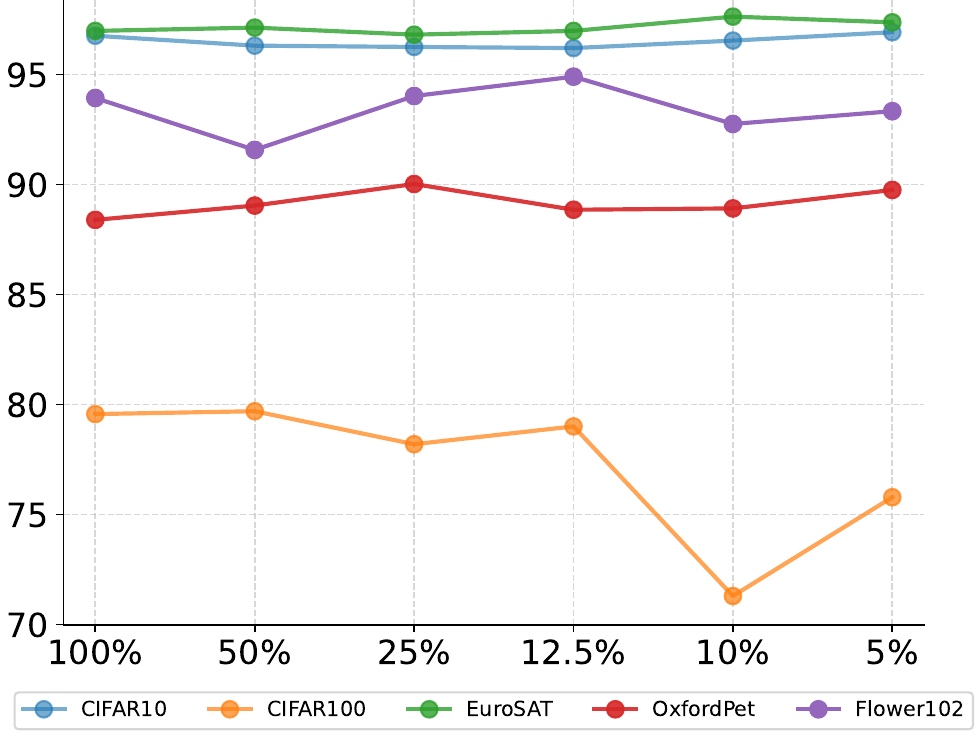}
    \caption{Dir-Global}
    \end{subfigure}
    \hfill
    \begin{subfigure}{0.28\linewidth}
    \includegraphics[width=\linewidth]{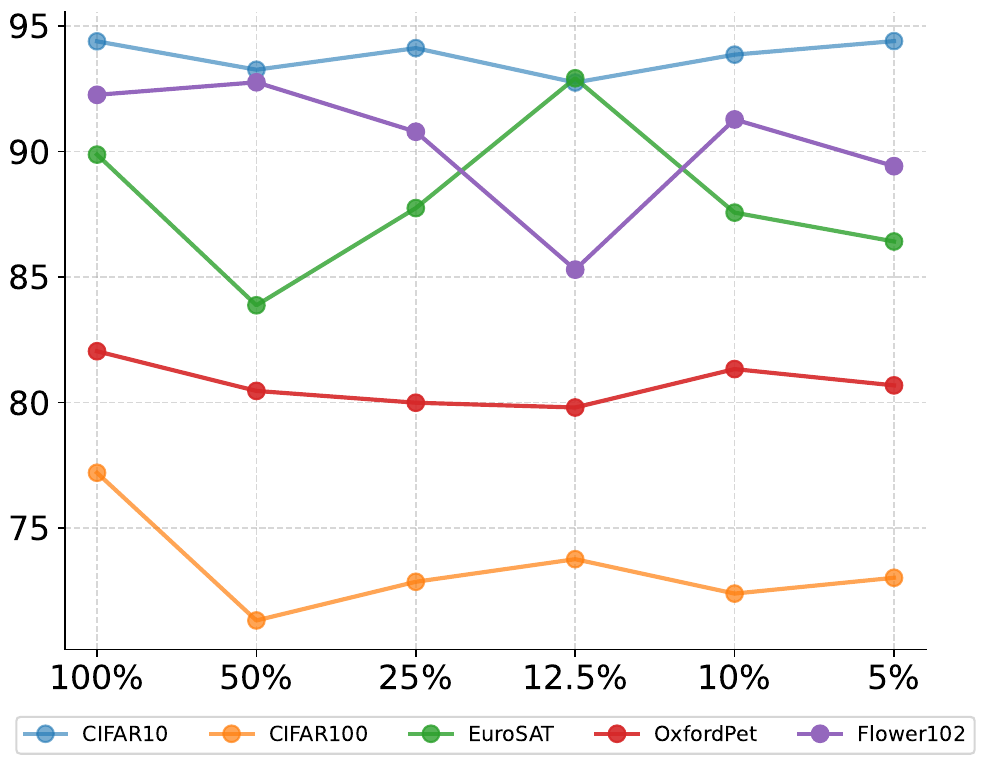}
    \caption{Path-Global}
    \end{subfigure} 
    \\
    \begin{subfigure}{0.28\linewidth}
    \includegraphics[width=\linewidth]{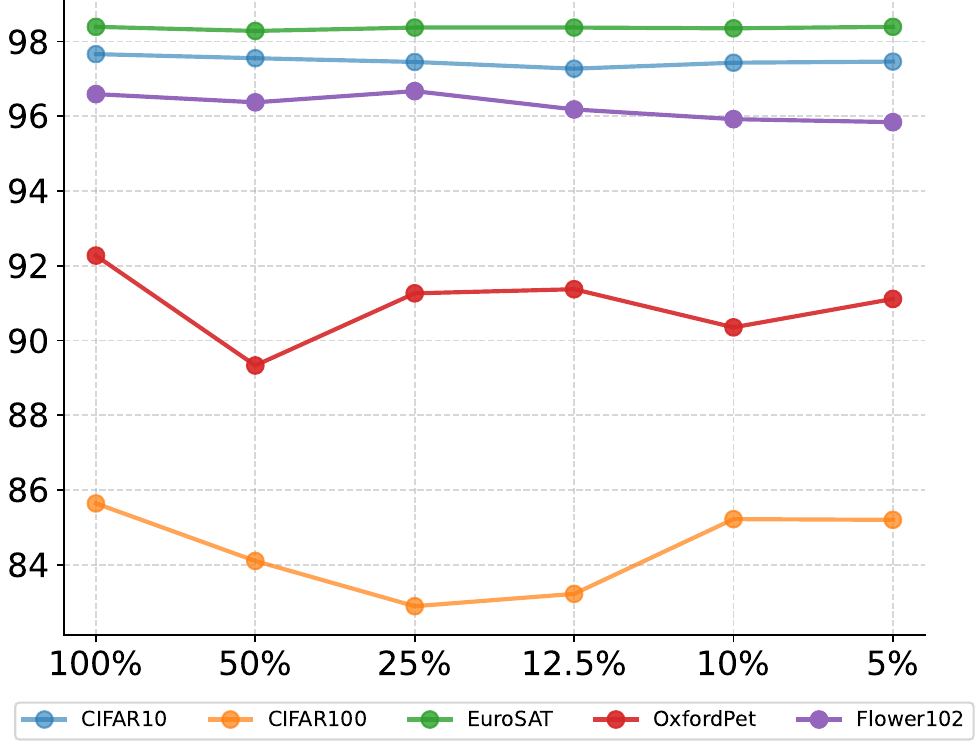}
    \caption{IID-Local}
    \end{subfigure}
    \hfill
    \begin{subfigure}{0.28\linewidth}
    \includegraphics[width=\linewidth]{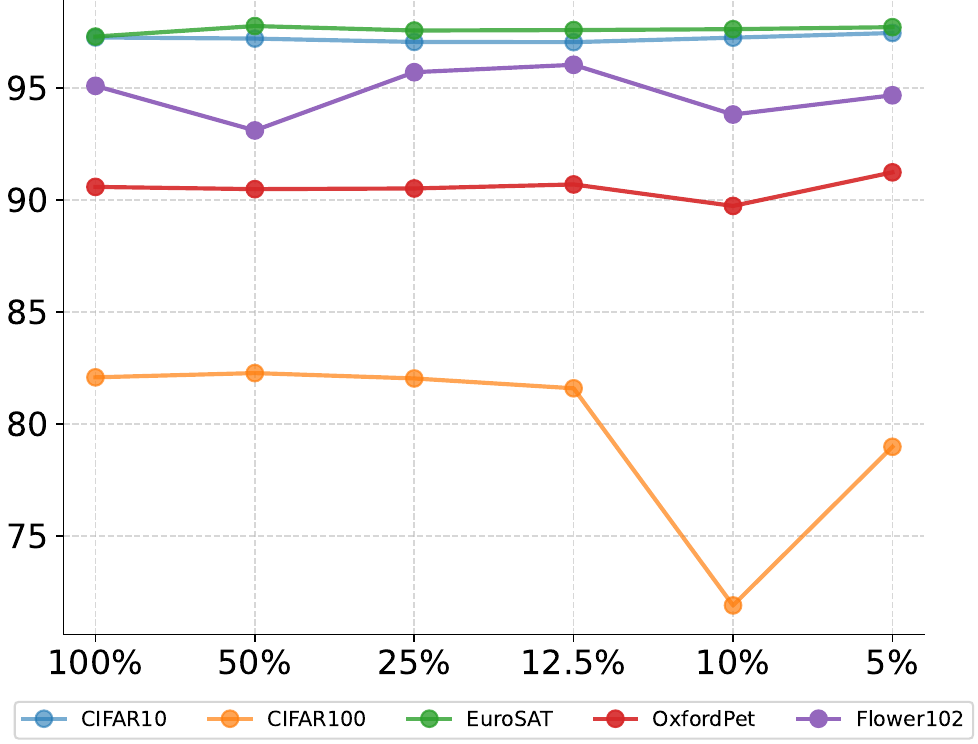}
    \caption{Dir-Local}
    \end{subfigure}
    \hfill
    \begin{subfigure}{0.28\linewidth}
    \includegraphics[width=\linewidth]{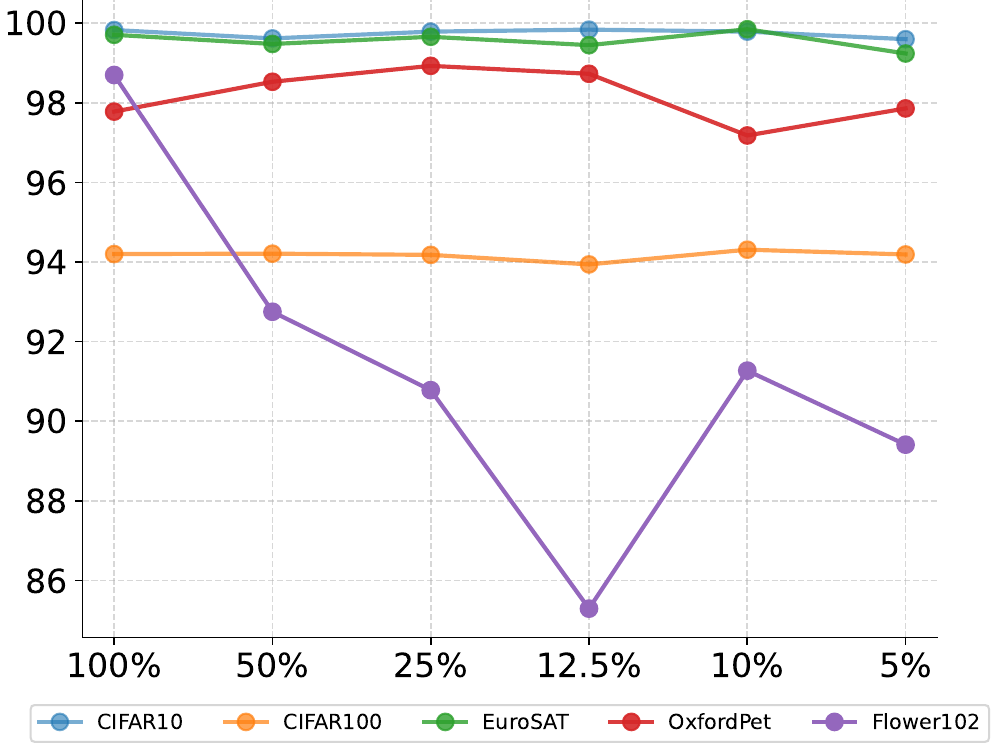}
    \caption{Path-Local}
    \end{subfigure} 
    \\
\caption{Global and average local test accuracy (\%) across different numbers of uploaded image features under different data settings with five benchmark datasets. The x-axis represents the uploaded proportion of image features.}
\label{fig:comm_accuracy_compare_proportion}
\end{figure*}

\begin{table}[!htb]
\centering
\resizebox{.47\textwidth}{!}{
    \begin{tabular}{l|ccccc}
    \toprule
    Setting & CIFAR10 & CIFAR100 & EuroSAT & OxfordPet & Flower102 \\
    \hline
    \rowcolor[rgb]{0.902,0.902,0.902} \multicolumn{6}{l}{Global Performance} \\
    IID   & $97.82^{+0.16}$ & $86.42^{+0.79}$ & $97.94^{-0.45}$ & $94.33^{+2.10}$ & $99.31^{+2.64}$  \\
    Dir   & $96.72^{-0.04}$ & $80.81^{+1.25}$ & $96.72^{-0.26}$ & $88.23^{-0.16}$ & $97.45^{+3.53}$  \\
    Path  & $95.40^{+1.02}$ & $76.18^{-1.02}$ & $91.63^{+1.76}$ & $75.91^{-6.13}$ & $92.25^{+0.00}$  \\
    \hline
    \rowcolor[rgb]{0.902,0.902,0.902} \multicolumn{6}{l}{Local Performance} \\
    IID   & $97.82^{+0.16}$ & $86.42^{+0.78}$ & $97.94^{-0.45}$ & $94.31^{+2.04}$ & $99.31^{+2.72}$   \\
    Dir   & $97.44^{+0.18}$ & $83.28^{+1.19}$ & $97.09^{-0.21}$ & $90.90^{+0.31}$ & $97.59^{+2.49}$  \\
    Path  & $99.66^{-0.17}$ & $93.41^{-0.79}$ & $99.81^{+0.10}$ & $98.90^{+1.12}$ & $99.70^{+1.00}$   \\
    \bottomrule
    \end{tabular}
}
\caption{Global and local performance (\%) with the pre-trained CLIP based on the Transformer architecture ViT-B/16 over five benchmark datasets under different data settings. The superscript values represent the accuracy difference with the architecture of ViT-B/32. ``+'' indicates that the performance on the ViT-B/16 architecture outperforms the ViT-B/32 architecture, while ``-'' denotes a performance decline.}
\label{tab:PM_comparison_ViT}
\end{table}

\begin{table*}[!htb]
\centering
\resizebox{.98\textwidth}{!}{
    \begin{tabular}{l|ccc||ccc||ccc}
    \toprule
    \multicolumn{1}{c|}{\multirow{2}{*}{Dataset}} & \multicolumn{3}{c||}{Cosine Similarity} & \multicolumn{3}{c||}{Dot Product} & \multicolumn{3}{c}{Euclidean Distance} \\
    \cline{2-10}
    \multicolumn{1}{c|}{} & IID   & Dir  & Path & IID   & Dir   & Path   & IID   & Dir   & Path \\
    \hline
    CIFAR10   & $97.66^{97.66}$ & $96.76^{97.26}$ & $94.38^{99.83}$ & $97.51^{97.51}$ & $96.78^{97.50}$  & $88.29^{99.79}$ & $97.28^{97.27}$ & $96.18^{96.63}$ & $91.17^{99.71}$  \\
    CIFAR100   & $85.63^{85.64}$ & $79.56^{82.09}$ & $77.20^{94.20}$ & $85.04^{85.05}$  & $78.50^{80.85}$ & $72.74^{94.35}$ & $85.13^{85.11}$ & $78.74^{80.28}$ & $76.55^{94.95}$ \\
    EuroSAT     & $98.39^{98.39}$ & $96.98^{97.30}$ & $89.87^{99.71}$ & $98.35^{98.36}$  & $97.19^{97.79}$ & $88.81^{99.45}$ & $98.19^{98.19}$ & $97.02^{97.42}$ & $83.98^{99.40}$  \\
    OxfordPet    & $92.23^{92.27}$ & $88.39^{90.59}$ & $82.04^{97.78}$ & $90.71^{90.69}$  & $86.67^{88.25}$ & $79.61^{96.29}$ & $90.60^{90.58}$ & $85.96^{85.84}$ & $73.04^{96.98}$  \\
    Flower102   & $96.67^{96.59}$ & $93.92^{95.10}$ & $92.25^{98.70}$ & $96.18^{96.18}$ & $94.71^{95.60}$ & $91.08^{98.90}$ & $97.55^{97.63}$ & $92.65^{93.22}$ & $92.45^{98.50}$ \\
    \bottomrule
    \end{tabular}
}
\caption{Global and local performance comparison (\%) of different similarity functions on five benchmark datasets under different data settings. The superscript values represent the average local performance among clients.}
\label{tab:similarity_function_comparision}
\end{table*}

\noindent \textbf{Aggregation Scheme for Local Inference.}
As illustrated in Figure~\ref{fig:local_aggregation_performance}, we evaluate the local performance under three data settings: \textit{IID}, \textit{Dir} with different $\alpha$, and \textit{Path}. 
We use `before' to represent the test accuracy on the local test set after the final round of local training and before the aggregation on the server side, while `after' denotes the test accuracy on the local test set after the parameter aggregation on the server.
We can observe that:
1) In \textit{IID} and \textit{Dir} data settings, the average local performance in FedAlign improves after the parameter aggregation. 
This is because clients have overlapping labels, allowing them to learn more knowledge from others via the aggregated parameters.
2) In contrast, the local performance declines after parameter aggregation in \textit{Path} data setting where clients have disjoint classes.
We speculate that the parameter aggregation may lead to a drift in the local optimal optimization direction, further affecting local performance. 
However, our aggregation method effectively adjusts the global optimization direction, preventing overfitting and improving global performance.
Consequently, we prefer to test local performance using local parameters of the final global round in the \textit{Path} setting, while using aggregated parameters in other data settings.
Notably, in the \textit{Path} setting, the global and local training are the same as the process in other data settings, only the inference for the local performance is different.

\noindent \textbf{Compared with Zero-shot CLIP.}
As shown in Table~\ref{tab:zero_shot_CLIP}, we display the test accuracy in zero-shot CLIP with pre-trained parameters.
Combined with the above experiment results, we find that:
1) FedOTP and FedPGP are sensitive to data heterogeneity, particularly FedOTP. 
They are easily influenced by local data distribution, causing the generalization capability degradation and performance reduction of models. 
2) The different behavior of test accuracy under the zero-shot CLIP with \textbf{GT} and \textbf{ST} input style reminds us to ensure the quality of text descriptions generated by LLM. 

\subsection{Details of Ablation Study}
\label{appendix_analysis_details}

\noindent \textbf{Hyperparameter Analysis.}
We also investigate the impact of four hyperparameters (\ie, loss factor $\mu$, the boundary layer of aggregation $m$, LoRA rank $r$, and the LoRA starting layer $l$) on local performance.
As shown in Figure~\ref{fig:hyper_parameter_compare_local}, we observe that:
1) The optimal hyperparameter for local performance aligns with those for global performance (see Figure 3 in the main text). 
Specifically, we set $\mu=0.1$, $m=9$, $r=4$, and $l=2$.
2) As illustrated in Figure~\ref{fig:boundary_layer_local}, the combined parameter aggregation scheme (\ie, the ex-query aggregation based on the query-based aggregation with a weighted aggregation) with the boundary layer $m=9$ improves local performance compared to the pure weighted aggregation scheme, especially in complex datasets.

\noindent \textbf{The Effect of Pre-trained Models.}
Then, we explore the effect of pre-trained model architectures on the model performance.
We conduct experiments using pre-trained CLIP based on the ViT-B/16 architecture, and the results are shown in Table~\ref{tab:PM_comparison_ViT}. 
The model performance with the ViT-B/32 and ViT-B/16 architecture is comparable on CIFAR10, CIFAR100, and EuroSAT datasets. 
However, the test accuracy in the ViT-B/16 architecture significantly outperforms the accuracy in the ViT-B/32 model over the Flower102 dataset. 
We hypothesize that the pre-trained model with a stronger generalization Transformer architecture can facilitate performance improvement, particularly on datasets with more complex images and diverse categories.
The generalization ability of different pre-trained models can be seen in Table~\ref{tab:zero_shot_CLIP}.
Furthermore, it is also important to consider the increased amount of trainable parameters, computational cost, and communication overhead with more complex pre-trained models.
For example, the trainable parameter count of an image encoder is $0.695$M in the CLIP over the ViT-B/16 architecture compared to $0.588$M in the CLIP over the ViT-B/32 architecture, and the corresponding amount is invariable in the text encoder. 

\noindent \textbf{Choice of Similarity Functions.}
As shown in Table~\ref{tab:similarity_function_comparision}, we present the performance comparison of FedAlign under three different similarity measurement functions:
\begin{itemize}
\item \textbf{Cosine Similarity}: measure the cosine of the angle between two vectors:
$$sim(x,y)=\frac{x \cdot y}{\|x\|\|y\|};$$
\item \textbf{Dot Product}: measure the product sum of the corresponding entries of the two vectors:
$$sim(x,y)=\sum_{i=1}^n{x_iy_i};$$
\item \textbf{Euclidean Distance}: measure the straight-line distance between two points:
$$sim(x,y)=\sqrt{\sum_{i=1}^n{(x_i-y_i)^2}}.$$
\end{itemize}
In both \textit{IID} and \textit{Dir} settings, the global and local model performance under three different functions are comparable. 
However, in the \textit{Path} setting, FedAlign with the cosine similarity function outperforms the other two functions. 
In other words, FedAlign demonstrates robustness over the similarity measurement function in \textit{IID} and \textit{Dir} settings, while in the \textit{Path} setting, we prefer to use the cosine similarity function.

\noindent \textbf{Impact of the Number of Uploaded Image Features.}
As illustrated in Figure~\ref{fig:comm_accuracy_compare_proportion}, we show the impact of different upload ratios of image features on model performance. 
The performance remains robust on the upload ratio under \textit{IID} and \textit{Dir} settings, and the upload ratio has a slight influence on performance under the \textit{Path} setting. 
Interestingly, a lower upload ratio cannot lead to performance degradation, which we attribute to the strong generalization of the pre-trained text encoder. 
Moreover, complex datasets are more sensitive to the upload ratios. 
These findings indicate that clients can reduce communication overhead while maintaining model performance.

\end{document}